\pdfoutput=1
\documentclass[11pt,final]{article}
\usepackage[preprint]{acl}
\usepackage{times}
\usepackage{latexsym}
\usepackage{booktabs}
\usepackage{float}
\usepackage[T1]{fontenc}
\usepackage[utf8]{inputenc}
\usepackage{microtype}
\usepackage{inconsolata}
\usepackage{graphicx}
\usepackage{listings}
\usepackage{xcolor}
\usepackage{arydshln}

\lstdefinestyle{customcpp}{
    language=C++,
    basicstyle=\ttfamily,
    keywordstyle=\color{blue},
    commentstyle=\color{gray},
    stringstyle=\color{red},
    breaklines=true,
    numbers=left,
    numberstyle=\tiny\color{gray},
    numbersep=5pt,
    showstringspaces=false
}

\lstdefinestyle{customsql}{
    language=SQL,
    basicstyle=\ttfamily,
    keywordstyle=\color{blue},
    commentstyle=\color{gray},
    stringstyle=\color{red},
    breaklines=true,
    numbers=left,
    numberstyle=\tiny\color{gray},
    numbersep=5pt,
    showstringspaces=false
}

\lstdefinestyle{customjava}{
    language=Java,
    basicstyle=\ttfamily,
    keywordstyle=\color{blue},
    commentstyle=\color{gray},
    stringstyle=\color{red},
    breaklines=true,
    numbers=left,
    numberstyle=\tiny\color{gray},
    numbersep=5pt,
    showstringspaces=false
}

\lstdefinestyle{customMarkdown}{
    language=Markdown,
    basicstyle=\ttfamily,
    keywordstyle=\color{blue},
    commentstyle=\color{gray},
    stringstyle=\color{red},
    breaklines=true,
    numbers=left,
    numberstyle=\tiny\color{gray},
    numbersep=5pt,
    showstringspaces=false
}

\lstdefinestyle{custompython}{
    language=Python,
    basicstyle=\ttfamily,
    keywordstyle=\color{blue},
    commentstyle=\color{gray},
    stringstyle=\color{red},
    breaklines=true,
    numbers=left,
    numberstyle=\tiny\color{gray},
    numbersep=5pt,
    showstringspaces=false
}
\lstdefinestyle{customjson}{
    language=Json,
    basicstyle=\ttfamily,
    keywordstyle=\color{blue},
    commentstyle=\color{gray},
    stringstyle=\color{red},
    breaklines=true,
    numbers=left,
    numberstyle=\tiny\color{gray},
    numbersep=5pt,
    showstringspaces=false
}

\lstdefinelanguage{markdown}{
    basicstyle=\ttfamily,
    sensitive=false,
    comment=[l]{\#},
    morecomment=[s]{<!--}{-->},
    commentstyle=\color{gray}\itshape,
    stringstyle=\color{red},
    moredelim=[is][\textbf]{**}{**},
    moredelim=[is][\textit]{*}{*},
    moredelim=[is][\texttt]{`}{`},
}

\lstdefinelanguage{json}{
    basicstyle=\ttfamily,
    numbers=left,
    numberstyle=\scriptsize,
    stepnumber=1,
    numbersep=8pt,
    showstringspaces=false,
    breaklines=true,
    frame=lines,
    backgroundcolor=\color{gray!10},
    literate=
     *{0}{{{\color{blue}0}}}{1}
      {1}{{{\color{blue}1}}}{1}
      {2}{{{\color{blue}2}}}{1}
      {3}{{{\color{blue}3}}}{1}
      {4}{{{\color{blue}4}}}{1}
      {5}{{{\color{blue}5}}}{1}
      {6}{{{\color{blue}6}}}{1}
      {7}{{{\color{blue}7}}}{1}
      {8}{{{\color{blue}8}}}{1}
      {9}{{{\color{blue}9}}}{1}
      {:}{{{\color{red}{:}}}}{1}
      {,}{{{\color{red}{,}}}}{1}
      {\{}{{{\color{red}{\{}}}}{1}
      {\}}{{{\color{red}{\}}}}}{1}
      {[}{{{\color{red}{[}}}}{1}
      {]}{{{\color{red}{]}}}}{1},
}

\usepackage{booktabs}  
\usepackage{fancybox}
\usepackage{framed}
\usepackage{mdframed}
\usepackage{afterpage}
\usepackage{adjustbox}
\usepackage{float}
\usepackage{enumitem}
\usepackage{hyperref}

\title{Infinite-Instruct: Synthesizing Scaling Code instruction Data with Bidirectional Synthesis and Static Verification}

\usepackage{fancyhdr}
\fancypagestyle{firstpage}{
  \fancyhf{}
  \fancyfoot[L]{\small $^{\dagger}$Work done at internship at ByteDance \quad $^{*}$Corresponding author  }

}
\author{
  \textbf{Wenjing Xing\textsuperscript{1}},
  \textbf{Wenke Lu\textsuperscript{1,2$^{\dagger}$}},
  \textbf{Yeheng Duan\textsuperscript{1}},
  \textbf{Bing Zhao\textsuperscript{1*}}, \\
  \textbf{Zhenghui Kang\textsuperscript{1}},
  \textbf{Yaolong Wang\textsuperscript{1}},
  \textbf{Kai Gao\textsuperscript{1}},
  \textbf{Lei Qiao\textsuperscript{1}}
\\
  \textsuperscript{1} ByteDance, \\
  \textsuperscript{2} Dalian University of Technology \\
  \small{
  \texttt{\{xingwenjing.417, duanyeheng, kangzhenghui, wangyaolong, gaokai, qiaolei\}@bytedance.com} 
  } \\
  \small{
  \texttt{luwenke@mail.dlut.edu.cn}
  }\\
  \small{
  \texttt{zhaobingcars@gmail.com}
  }
}

\begin{document}
\maketitle
  
\begin{abstract}
Traditional code instruction data synthesis methods suffer from limited diversity and poor logic. We introduce Infinite-Instruct, an automated framework for synthesizing high-quality question-answer pairs, designed to enhance the code generation capabilities of large language models (LLMs). The framework focuses on improving the internal logic of synthesized problems and the quality of synthesized code. First, "Reverse Construction" transforms code snippets into diverse programming problems. Then, through "Backfeeding Construction," keywords in programming problems are structured into a knowledge graph to reconstruct them into programming problems with stronger internal logic. Finally, a cross-lingual static code analysis pipeline filters invalid samples to ensure data quality. Experiments show that on mainstream code generation benchmarks, our fine-tuned models achieve an average performance improvement of 21.70\% on 7B-parameter models and 36.95\% on 32B-parameter models. Using less than one-tenth of the instruction fine-tuning data, we achieved performance comparable to the Qwen-2.5-Coder-Instruct. Infinite-Instruct provides a scalable solution for LLM training in programming. We open-source the datasets used in the experiments, including both unfiltered versions and  filtered versions via static analysis. The data are available at ~\url{https://github.com/xingwenjing417/Infinite-Instruct-dataset}
\end{abstract}

\section{Introduction}
Recent advances in Large Language Models (LLMs) have revolutionized code generation capabilities through instruction tuning\citep{Zhang:2023}. The effectiveness of this paradigm fundamentally depends on high-quality instruction-response pairs that enable models to comprehend and execute diverse programming tasks\citep{Chen:2021}. However, scaling the acquisition of such instruction data presents a significant bottleneck\citep{Wang:2024}, as conventional manual annotation methods are both resource-intensive and constrained in their coverage.

Prior research has explored various approaches to automatic instruction synthesis. Early attempts leveraging Self-Instruct\citep{Wang:2022} demonstrated the potential of utilizing existing LLMs in a teacher-student framework for instruction synthesis. Despite their initial success, these approaches exhibited inherent limitations due to biases inherited from both the foundation models and the limited seed examples\citep{Yu:2023}. While subsequent developments like Evol-Instruct\citep{Luo:2023}\citep{Taori:2023} introduced more sophisticated instruction evolution mechanisms, they remained bounded by predetermined evolution rules. OSS-Instruct\citep{wei:2023} achieved a significant advancement by pioneering the utilization of real-world code snippets as inspiration sources. However, this approach lacked robust quality assurance mechanisms and structural validation protocols.

To address these limitations, we present Infinite-Instruct, a novel bidirectional framework for code-oriented data synthesis that extends the capabilities of OSS-Instruct. The framework operates through two complementary mechanisms: Reverse Construction and Backfeeding Construction. The Reverse Construction component transforms diverse code fragments into comprehensive programming tasks by analyzing code structure and functionality. The Backfeeding mechanism establishes vocabularies of tasks, instructions, and knowledge points through a complex knowledge graph architecture, transforming phrases into programming tasks. This bidirectional approach enables the framework to infinitely utilize code fragments and maintained vocabularies, continuously leveraging the model to generate unlimited high-quality training data.

Our methodology incorporates four key innovations with specific implementation details: (1) a bidirectional code-prompt evolution framework that combines structural code analysis with semantic understanding. (2) a knowledge-graph-enhanced backfeeding mechanism that creates a closed-loop optimization  by extracting and refining keywords from synthetic prompts. 
(3) a comprehensive seven-dimensional quality assessment protocol implemented through a cross-language static code analysis pipeline, which effectively filters invalid samples and ensures instruction validity. (4) systematic knowledge integration through structured vocabulary tables that maintain consistent terminology and conceptual relationships. 

\section{Related Work}

\subsection{Advances in Automated Instruction Synthesis} 
Automated instruction  synthesis has made breakthroughs in recent years, contributing to the development of open-source models, represented by Nemotron-4\citep{Adler:2024}. \citep{Wang:2023} pioneered Self-Instruct, which synthesizes instruction prompts by bootstrapping from a small seed set of human-written examples. This method allowed LLMs to generate both new instructions and corresponding instances, reducing dependence on manual annotation. However, Self-Instruct's synthesized prompts often lacked complexity and diversity compared to expert-created instructions. \citep{Xu:2023} addressed this limitation with Evol-Instruct in their work WizardLM , introducing evolutionary synthesis method that systematically increase complexity through operations like adding constraints, deepening context, and increasing reasoning steps. While Evol-Instruct successfully generated more challenging prompts, it operated primarily in a direction from code snippets  to prompts, evolving existing instructions without systematically creating novel prompt types. 

For code domains, \citep{Luo:2023} developed WizardCoder, adapting evolutionary instruction synthesis specifically for programming tasks with code-specific constraints and debugging scenarios. \citep{wei:2023} proposed OSS-INSTRUCT, which uniquely leverages open-source code snippets as inspiration for generating diverse coding problems, mitigating LLM bias through real-world code references. Despite these advances, existing prompt synthesis approaches remain limited by their evolution patterns, insufficient coverage across knowledge domains, limited quality assurance beyond basic validity checks, and lack of closed-loop optimization mechanisms.

\subsection{Static Code Analysis}
Static code analysis is a method that identifies potential issues without executing the program, with linter systems being its powerful implementation. These systems support multiple programming languages including JavaScript, Python, Ruby, C/C++, and Java, detecting syntax errors, code smells, and style deviations, while sometimes identifying performance bottlenecks and security vulnerabilities. 

In recent years, with the increasingly widespread application of large language models (LLMs) in code generation and comprehension, static code analysis has evolved to evaluate and enhance the quality of AI-generated code. For instance, the Llama 3 series~\citep{grattafiori:2024} employs parsers and linters to guarantee syntactic accuracy in all generated code, detecting errors such as syntax mistakes, use of uninitialized variables, non-imported functions, code style issues, and typing errors. Similarly, Qwen2.5-Coder~\citep{Hui:2024} incorporates abstract syntax tree parsing to filter out code snippets containing parsing errors. 
It indicates that LLMs for code generation are increasingly integrating multilingual static code analysis capabilities during their datas construction phase to ensure code quality across multiple programming languages.

\section{Method}
\begin{figure*}[t]
\centering
\includegraphics[width=0.8\paperwidth]{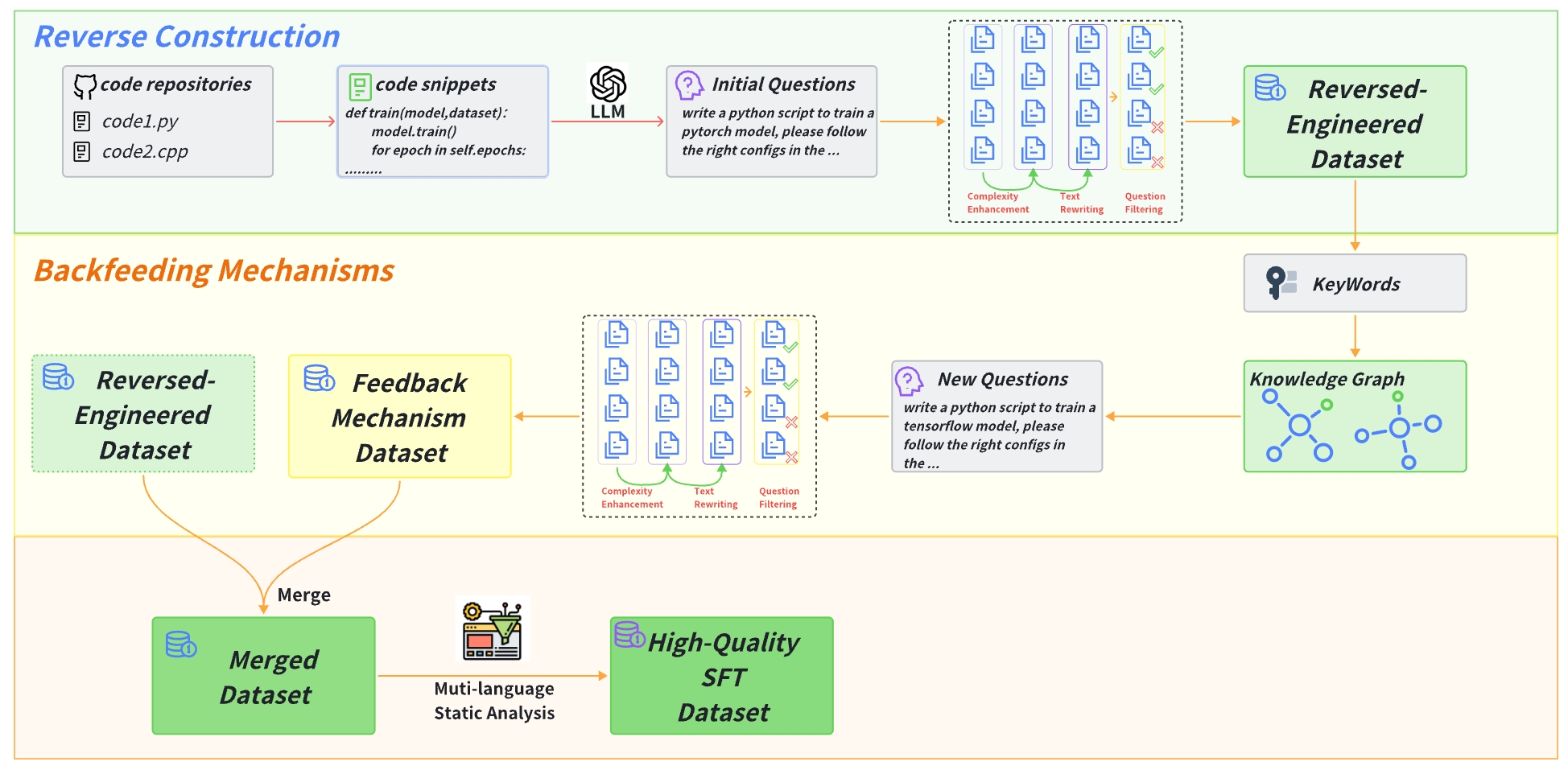} 
\caption{\label{fig:Automated Prompt Synthesis Technology Roadmap}Automated Prompt Synthesis Technology Roadmap.}
\end{figure*}
We propose Infinite-Instruct, a code-oriented bi-directional instruction synthesis method that generates high-quality and diverse instruction data. It revolves the inter-evolution between code and problems, forming a closed-loop optimization generation system through the reverse construction (“Code → Problem”)  and the backfeeding construction (“Keyword → Problem”) .The complete data synthesis process can be seen in ~\autoref{fig:Automated Prompt Synthesis Technology Roadmap}.

During Reverse Construction, we leverage a large language model (LLM) to analyze numerous code snippets and generate diverse programming problems. These problems undergo complexity adjustment and textual rewriting to ensure quality and variety. During BackFeeding Construction, a knowledge graph is established to present the semantic relationships among keywords extracted from the synthesized problems. At last, the generated instruction data is evaluated and filtered by a seven-dimensional quality assessment framework.

\subsection{Reverse Construction: Deriving Problems from Code}
We extract multilingual code from Magicoder-OSS-Instruct-75K \citep{wei:2023} as our seed repository,  it reduces the need of data cleaning. We randomly extract code snippets of 5-20 lines multiple times, ensuring these snippets maintain functional completeness and independence after deduplication.

We design specialized prompt templates for different types of problems (see ~\ref{Integrated Reverse Question Generation Template} for details). By analyzing the characteristics of each code snippet, we guide GPT-4o-0806 to consider possible application scenarios, identify valuable problem points, and improve question quality accordingly. To increase the difficulty of questions, we rewrite problems from six directions, including adding constraints, depth extension, concretization, reasoning refinement, input enhancement, and innovative changes(see ~\ref{Complicate Prompt}). It formulates questions with richer constraints and contextual information that more closely resemble real-world application scenarios.

We also discover that synthesized questions usually lack expression diversity (e.g., most code generation questions begin with "Please help me write a..."). To address this issue, we rewrite the text by analyzing sentence logical relationships and restructuring paragraph structures, making the questions more aligned with real users' questioning styles(refer to ~\ref{Text Rewrite Prompt}).

Through this process, We synthesize approximately 30K problems. The advantage of this approach is that it breaks through the limitations of predefined datasets and can generate more diverse questions. At the same time, the generated questions are closely related to the code, ensuring their solvability.

\subsection{Backfeeding Construction: Closed-Loop Optimization from Keywords to Problems}
While generating programming problems based on randomized code snippets yields diversity, it often lacks logical structure and pedagogical focus. 

We define three types of keywords for a programming problem(tasks, instructions, and knowledge points). Tasks and instructions are extracted from "reverse" synthesis problems (refer to ~\ref{Prompts for keyword extraction}), and after deduplication, we construct a vocabulary containing keywords from all three categories and generate random combinations. Initial knowledge points sourced from online programming tutorials (like "Runoob Tutorial") and official SQL documentation (including MySQL, Hive, SQLite, PostgreSQL, Oracle, etc.), supplements with secondary tags from the field of computer science. Each programming language has approximately 10K words per dimension on average, with about 20K secondary tags collected from various fields of computer science (such as algorithms, software engineering, etc.).

To improve the effectiveness of the problems, we use the concept of knowledge graphs to construct prompts as follows: First, we extract entities from the keywords and categorize them into tasks, instructions, and knowledge points, forming a list of nodes. Second, we analyze the relationships between these nodes and construct triples to represent the connections between them, then extract keyword combinations with clear semantic associations and rigorous logical structures from these triples. We deduplicate the final keyword groups using cosine similarity (threshold 0.8) and use these groups to synthesize seven types of "backfeed" problems, including Code Generation, Code Understanding, Knowledge-based Question, Code Completion, Code Optimization, Debug, and Modify Code as required. We use GPT-4o-0806 model as the engine to drive the entire process, ultimately synthesize approximately 20K instructions. The prompt template for "backfeed" problem synthesis can be found in ~\ref{Keyword combination and filtering} and ~\ref{Integrated Backfeeding Question Generation Template}.

\subsection{Prompt Filtering}
For both “Reverse” and “Backfeeding” construction, we introduce seven quality metrics of Arena-Hard-Auto~\citep{li:2024} to filter out high-quality question, including the metrics of specificity, domain knowledge, complexity, problem solving ability, creativity, technical accuracy, and practical application. We use GPT-4o-0806 to evaluate the questions that the system generates. For each question, we conduct three assessments, and the final score is the average of the three assessments. We filter out data with scores less than 6 points. See ~\ref{Prompt Filtering} for specific cue words. Finally, we filter out 10K “reverse” data and 10K “backfeed” data.

\begin{figure*}[t]
\centering
\includegraphics[width=0.8\paperwidth]{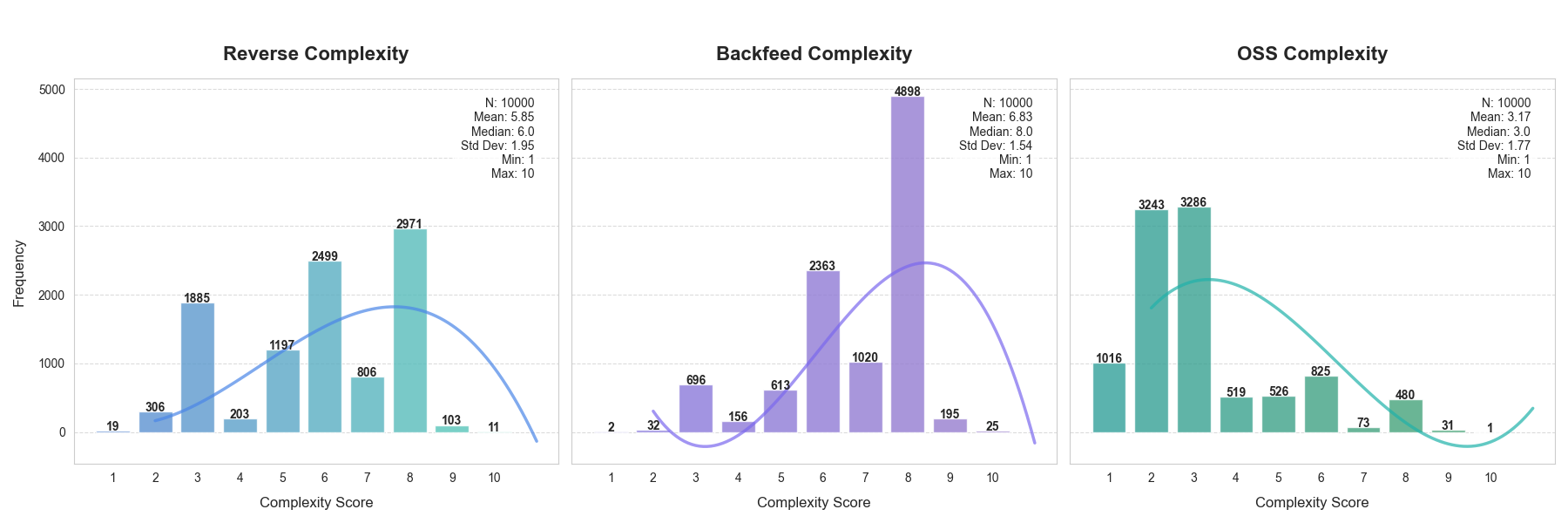} 
\caption{\label{fig:Complexity Score Comparison}Complexity Score Comparison}
\end{figure*}

Three datasets are included in our analysis: Reverse, Backfeed, and OSS-Instruct. Each dataset is assessed using a complexity score that ranges from 1 to 10.A detailed explanation of the scoring system can be found in ~\ref{Instruction Complexity Assessment}. To ensure consistency, 10,000 samples are randomly selected from each dataset for evaluation. As shown in ~\autoref{fig:Complexity Score Comparison}, the results indicate:

\textbf{Backfeed}: highest complexity (mean 6.83, median 8.0), standard deviation 1.54, difficulty is more concentrated at high levels.

\textbf{OSS-Instruct}: lowest complexity (mean 3.17, median 3.0), suggesting that open source instructions are generally simpler and more straightforward.

\textbf{Reverse}: moderate complexity (mean 5.85, median 6.0), largest standard deviation (1.95), a more balanced distribution of instruction difficulty.

These differences are important guidelines for training strategies: Backfeed is suitable for training advanced reasoning ability, OSS is suitable for basic function training, and Reverse is suitable for comprehensive training due to its balanced characteristics. Taken together, the complexity differences of different datasets can provide guidance for hybrid training strategies, choosing the appropriate ratio of dataset combinations according to the needs of different application scenarios, such as mixing Backfeed and Reverse to optimize model performance.

\subsection{Response Quality Enhancement Based on Static Syntax Analysis}
We generate corresponding responses using GPT-4o-0816 for the above prompts. To ensure the quality of generated data, we employ language-specific static analysis tools as shown in ~\autoref{tab:static-tools}. Each tool is configured with customized rule settings categorized into three levels: disabled, error, and info. Disabled rules represent unnecessary checks based on our experience (e.g., @typescript-eslint/quotes in ESLint, which enforces consistent quote style). Error-level rules identify critical issues according to each linter's severity classifications, while remaining checks are set to info level. When static analysis detects error-level issues in generated code, we remove the entire prompt-response pair from our dataset rather than attempting to fix them automatically. This strict filtering approach ensures that only syntactically valid and high-quality code samples remain in our final dataset, while still accommodating acceptable stylistic variations flagged at the info level.

\begin{table}[H]
  \centering
  \begin{tabular}{lc}
    \hline
    \textbf{Programming Language} & \textbf{Static Analysis Tool} \\
    \hline
    Python & PyLint \\
    JavaScript & ESLint \\
    Java & Checkstyle \\
    C/C++ & Clang-Tidy \\
    SQL & SQLFluff \\
    \hline
  \end{tabular}
  \caption{Static analysis tools used for different programming languages}
  \label{tab:static-tools}
\end{table}

Our screening identifies a substantial number of responses with syntax errors—approximately 10,000 out of 100,000 generated responses contained syntax issues, including incorrect indentation, non-code content mixed within code blocks, and erroneous class instantiation methods (detailed cases are provided in ~\autoref{Static Analysis Examples}).

\section{Experiments}

\subsection{Experimental Settings}
We select Qwen2.5-Coder-7B-Base and Qwen2.5-Coder-32B-Base models as baseline to investigate the effectiveness of our proposed dataset. We transfer the SFT data to the same type of schema, which is a one-prompt and one-response format. We set the loss mask parameter of prompt to 0.0 and the response loss mask to 1.0. When training, we use the Mariana \citep{zou:2014} platform to implement the SFT training stage. Meanwhile,we make the training process parallelized on 4 machines,each equipped with 8 Nvidia A100-SXM-80GB GPUs. We set the global batchsize to a fixed integer 512, and run for 3 global epochs for each model to reach the endpoint. Also,we set the initial learning rate at $2 \times 10^{-5}$,and simultaneously we set the learning rate to decay in a cosine annealing manner. 

In the evaluation phase, we adapt a greedy generation strategy, specifically setting the temperature to 0 and the Top-K value to 1, ensuring deterministic outputs during each evaluation. For each evaluation instance across all models, we perform exactly one sampling. 

We prepare multiple datasets as follows. The original Magicoder-OSS-Instruct-75K dataset (generated by ChatGPT 3.5 turbo 1106) and its distilled version using the newer GPT-4o-0806 model, Magicoder-OSS-Instruct-75K-GPT-4o (to eliminate performance differences caused by model upgrades).Reverse-100K, a dataset of 100,000 coding task instances that we create using our reverse construction method, and its filtered version Reverse-90K that contains 90,000 instances after static code analysis. Backfeed-100K, a 100,000-instance coding dataset that we build through our backfeeding method, and its filtered version Backfeed-90K that contains 90,000 instances.

We select the following benchmarks to evaluate the coding ability of the models, MultiPL-E HumanEval \citep{cassano:2023}, MBPP \citep{austin:2021},MBPP+  \citep{liu:2023b}, Aider,BigCodeBench \citep{zhuo:2024}, LiveCodeBench \citep{jain:2024}. These benchmarks include mostly all the mainstream programming languages and cover a wide range of coding problems from easy to hard. 

At the same time, in order to observe the impact on general problem solving abilities while monitoring changes in the model's code capabilities, we use  Arena hard \citep{li:2024}. , a high-quality and reliable benchmark to test our trained models.

\begin{table*}[!ht]
\caption{Experimental Results Based on Qwen2.5-Coder-7B-Base}
\label{Experimental Results Based on Qwen2.5-Coder-Base-7B}
\centering
\footnotesize
\setlength{\tabcolsep}{4pt} 
\renewcommand{\arraystretch}{1.2}
\begin{tabular}{lccccccc}
\toprule
Model & MBPP & MBPP plus & MPL-E Human & Aider & BigCode Bench & LiveCode Bench & Arena Hard \\
\midrule
OSS-75K & 74.80 & 70.63 & 61.93 & 43.61 & 42.02 & 32.12 & 8.96 \\
OSS-75K-GPT-4o & 73.40 & 70.90 & 64.77 & 42.86 & 48.42 & 34.50 & 42.85 \\
Reverse 100K & 76.40 & 72.49 & 67.05 & 44.36 & 49.74 & 32.82 & 48.09 \\
Reverse 90K & 75.80 & 71.69 & 67.28 & \textbf{45.11} & 50.61 & 34.22 & 47.76 \\
Backfeed 100K & 71.60 & 67.99 & 55.24 & 42.11 & 50.61 & 35.34 & 47.70 \\
Backfeed 90K & 75.80 & 73.54 & 65.80 & 41.35 & 50.09 & 34.08 & 49.56 \\
Rev+Back 200K & 76.60 & 72.75 & 67.22 & 42.11 & 50.79 & 36.47 & 46.94 \\
Rev+Back 180K & \textbf{77.80} & \textbf{73.81} & \textbf{67.35} & 42.86 & \textbf{50.96} & \textbf{37.03} & \textbf{49.59} \\
\midrule
Qwen-2.5-Coder-7B-Instruct & 83.50 & 71.7 & 76.5 & 55.6 & 41.0 & 18.2 & - \\
\bottomrule
\end{tabular}
\end{table*}

\begin{table*}[!ht]
\caption{Experimental Results Based on Qwen2.5-Coder-32B-Base}
\label{Experimental Results Based on Qwen2.5-Coder-Base-32B}
\centering
\footnotesize
\setlength{\tabcolsep}{4pt} % 增加列间距，因为现在有更多空间
\renewcommand{\arraystretch}{1.2}
\begin{tabular}{lccccccc}
\toprule
Model & MBPP & MBPP plus & MPL-E Human & Aider & BigCode Bench & LiveCode Bench & Arena Hard \\
\midrule
OSS-75K & 81.40 & 74.34 & 67.58 & 48.87 & 52.46 & 34.22 & 19.71 \\
OSS-75K-GPT-4o & 82.60 & 76.98 & 76.12 & 57.14 & 55.35 & 50.49 & 67.40 \\
Reverse 100K & 83.20 & 76.46 & 76.48 & 57.89 & 57.28 & 49.79 & 66.95 \\
Reverse 90K & \textbf{83.40} & 75.13 & \textbf{76.65} & \textbf{59.40} & \textbf{57.72} & 49.09 & 67.38 \\
Backfeed 100K & 80.80 & \textbf{76.98} & 76.08 & 57.89 & 56.93 & \textbf{51.19} & 68.23 \\
Backfeed 90K & 81.20 & 75.40 & 76.55 & 57.14 & 56.84 & \textbf{51.19} & 67.93 \\
Rev+Back 200K & 82.40 & 76.46 & 76.55 & 56.39 & 56.67 & 50.49 & \textbf{70.17} \\
Rev+Back 180K & 83.20 & 75.13 & \textbf{76.65} & 56.39 & 56.32 & 49.51 & 67.97 \\
\midrule
Qwen-2.5-Coder-32B-Instruct & 90.2 & 75.1 & 79.4 & 60.9 & 49.6 & 31.4 & - \\
\bottomrule
\end{tabular}
\scriptsize
\begin{flushleft}
Note: MPL-E Human = MultiPLE-E Humaneval.
Rev+Back 200K = Reverse 100K + Backfeed 100K.
Rev+Back 180K = Reverse 90K + Backfeed 90K.
Bolded numbers represent the highest scores of all models.
\end{flushleft}
\end{table*}

\subsection{Experimental Results and Analysis}

\subsubsection{Improvements in Code Generation Capabilities}

The experimental results reveal several key insights regarding the effectiveness of different data construction and enhancement strategies.

\textbf{Effectiveness of the Reverse Construction.} 
~\autoref{Experimental Results Based on Qwen2.5-Coder-Base-7B} demonstrates that, for the 7B model, the Reverse Construction method significantly improves performance across all benchmarks. Reverse 90K achieves the highest score of 45.11 on the Aider dataset, and both Reverse 90K and Reverse 100K outperform the baseline OSS-75K, with average improvements of 18.81\% and 18.37\%, respectively.~\autoref{Experimental Results Based on Qwen2.5-Coder-Base-32B} highlights similar trends for the 32B model, where Reverse 90K achieves state-of-the-art results on MPL-E Human (76.65), Aider (59.40), BigCode Bench (57.72), and Arena Hard (67.38). The method delivers an average improvement of 36.64\% over OSS-75K, further demonstrating its effectiveness in enhancing model performance across diverse evaluation datasets.

\textbf{Effectiveness of the Backfeeding Construction.} 
On the Qwen2.5-Coder-7B-Base model, as shown in ~\autoref{Experimental Results Based on Qwen2.5-Coder-Base-7B}, the Backfeed strategy shows notable improvements over the OSS-75K.For the Qwen2.5-Coder-32B-Base model, Backfeed 100K also demonstrates significant advantages, as seen in ~\autoref{Experimental Results Based on Qwen2.5-Coder-Base-32B}. It achieves state-of-the-art results on multiple datasets, including MBPP plus (76.98) and LiveCode Bench (51.19), as shown in ~\autoref{Experimental Results Based on Qwen2.5-Coder-Base-32B}. Notably, it outperforms the larger Qwen2.5-Coder-32B-Instruct model on BigCode Bench (56.93 vs. 49.6) and achieves comparable performance on Aider (57.89 vs. 60.9), highlighting its efficiency and robustness even with smaller training datasets. These results demonstrate the scalability and versatility of Backfeed Construction across diverse evaluation scenarios.

\textbf{Superior Performance of Combined Strategies.}
The combination of Reverse 90K + Backfeed 90K performs best on the 7B model. ~\autoref{Experimental Results Based on Qwen2.5-Coder-Base-7B} shows that the combined strategy after static analysis surpasses Qwen2.5-Coder-7B-Instruct on MBPP plus, BigCode Bench, LiveCode Bench, and Arena Hard, making it the strongest performing model among all. On other evaluation sets, the combined strategy also demonstrates better scores than OSS-75K-GPT-4o.This is sufficient to demonstrate that combined strategies exhibit stable performance across all tests, reducing the volatility associated with single strategies.

\begin{figure}
\centering
\includegraphics[width=1.1\linewidth]{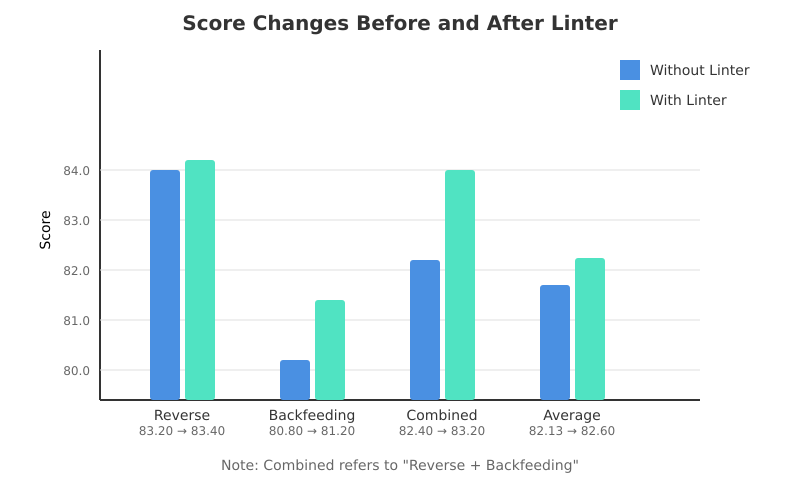}
\caption{\label{fig:Benchmark with MBPP}Benchmark with MBPP}
\end{figure}

\begin{figure}
\centering
\includegraphics[width=1.1\linewidth]{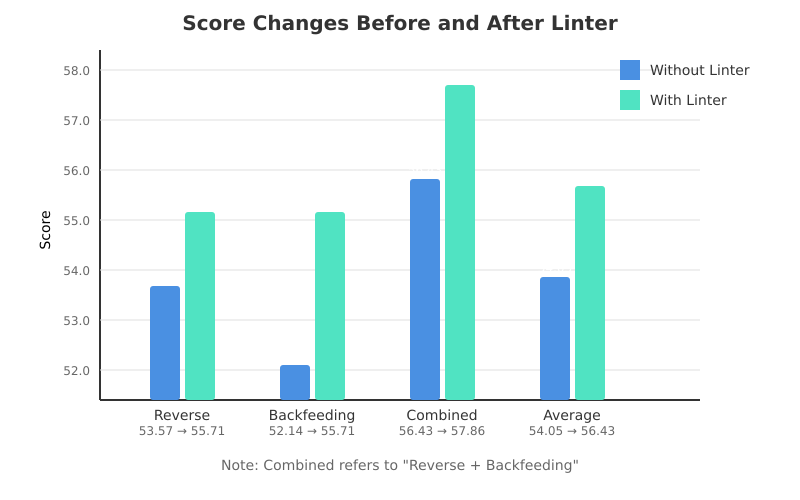}
\caption{\label{fig:Benchmark with MHPP}Benchmark with MHPP}
\end{figure}

\textbf{Impact of Static Analysis Filtering.} 
As can be seen from ~\autoref{tab:4，baseline-comparison-OSS-Instruct} and ~\autoref{tab:5，baseline-comparison-OSS-75K-GPT-4o},Datasets filtered through syntax checking (linter) generally perform better than their unfiltered versions. On the 7B model, Backfeed 90K shows significantly greater improvement than Backfeed 100K (19.30\% vs 8.75\%). On the 32B model, the trends in performance are similar(9.56\% vs -1.54\%).It demonstrates the important influence of high-quality instruction data.Based on the statistics from the MBPP and MBPP+ evaluation datasets, data that has undergone static syntax checking tends to achieve higher scores than data that has not been checked. The specific performance improvement effects can be referred to in~\autoref{fig:Benchmark with MBPP} and ~\autoref{fig:Benchmark with MHPP}.

\begin{table}[htbp]
\centering
\caption{Average Performance Improvement Compared to the original OSS-Instruct dataset}
\label{tab:4，baseline-comparison-OSS-Instruct}
\footnotesize
\setlength{\tabcolsep}{1.5pt}
\renewcommand{\arraystretch}{1.2}
\begin{tabular}{@{}lcc@{}}
\toprule
Model & Based on & Based on \\
 & Qwen2.5-coder-7B & Qwen2.5-coder-32B \\
\midrule
Reverse 100K & 18.37\% & 33.66\% \\
Reverse 90K & 18.81\% & 36.64\% \\
Backfeed 100K & 8.75\% & 32.84\% \\
Backfeed 90K & 19.30\% & \textbf{36.95\%} \\
Rev+Back 200K & 21.17\% & 36.82\% \\
Rev+Back 180K & \textbf{21.70\%} & 34.95\% \\
\bottomrule
\end{tabular}

\vspace{18pt}

\caption{Average Performance Improvement Compared to OSS-75K-GPT-4o dataset}
\label{tab:5，baseline-comparison-OSS-75K-GPT-4o}
\footnotesize
\setlength{\tabcolsep}{1.5pt}
\renewcommand{\arraystretch}{1.2}
\begin{tabular}{@{}lcc@{}}
\toprule
Model & Based on & Based on \\
 & Qwen2.5-coder-7B & Qwen2.5-coder-32B \\
\midrule
Reverse 100K & 7.62\% & 1.48\% \\
Reverse 90K & 9.59\% & \textbf{2.71\%} \\
Backfeed 100K & -1.54\% & 0.91\% \\
Backfeed 90K & 9.56\% & 2.59\% \\
Rev+Back 200K & \textbf{12.45\%} & 2.62\% \\
Rev+Back 180K & 11.92\% & 2.18\% \\
\bottomrule
\end{tabular}

\raggedright
\scriptsize
Note: \\
Rev+Back 200K = Reverse 100K + Backfeed 100K\\
Rev+Back 180K = Reverse 90K + Backfeed 90K\\
Bolded numbers represent the highest scores of all models.
\end{table}

\subsubsection{Comparative Analysis with Benchmark Methods}
\begin{figure}[H]
\centering
\includegraphics[width=1\linewidth]{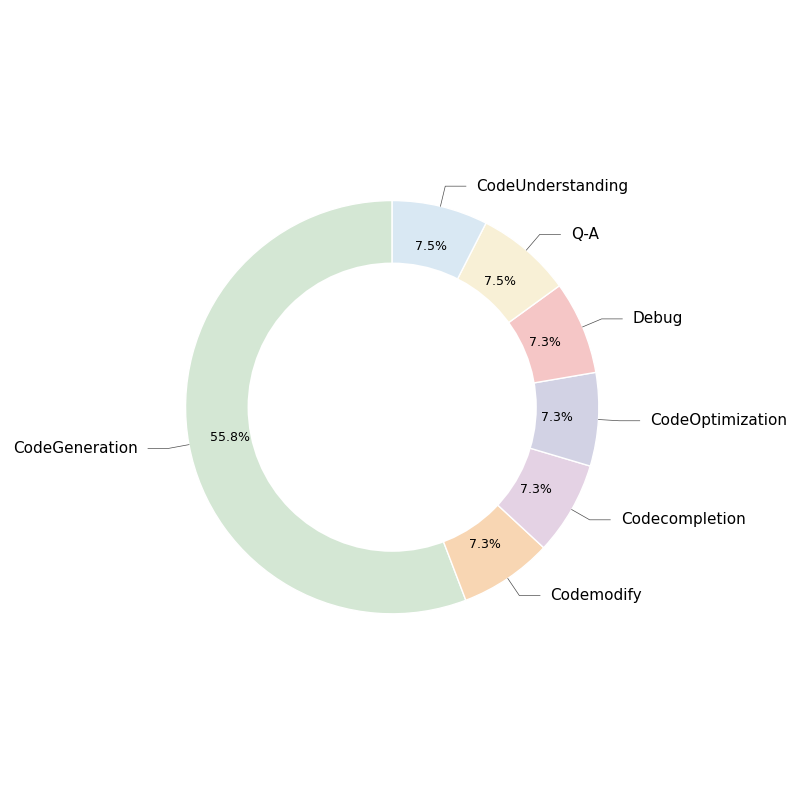}
\caption{\label{fig:Percentage of topic types}Percentage of topic types}
\end{figure}

\textbf{Enhanced Diversity.} Our instruction data covers various code-related tasks, as shown in ~\autoref{fig:Percentage of topic types}. The diversities enable stable model performance across different evaluation sets, especially in the complex MultiPle-EHumaneval.
\textbf{Quality Improvement.} Magicoder-OSS-Instruct-75K-GPT-4o, shows improvements over the original Magicoder-OSS-Instruct-75K dataset in most benchmarks. 
It indicates that as the performance of the LLM used to generate responses improves, the quality of the dataset tends to also increase. Datasets filtered through static analysis (Reverse 90K and Backfeed 90K) further enhance performance, indicating that code quality filtering is crucial for training effectiveness.

\textbf{Greater Efficiency.}
According to the official report released by the Qwen team~\citep{Hui:2024}, the Qwen2.5-coder-instruct model was fune-tuned on Qwen2.5-coder-Base model with millions of high-quality prompts. After that, they also employed mixed tuning and Direct Preference Optimization (DPO) during post-training. In contrast to the millions of instruction samples and complex post-training methods utilized by the Qwen team, our approach achieve comparable or even superior performance using at most 200K data samples. This outcome further validates the effectiveness of our data construction strategies.
Our synthetic strategies perform exceptionally well on the 32B model, outperforming the Qwen2.5-coder-instruct model in multiple tests. In the MBPP+ test, the Backfeed 100K strategy enable the 32B model to reach 76.98, higher than Qwen2.5-coder-instruct's 75.1. On BigCodeBench and LiveCodebench, all synthetic data strategies achieve evaluation results superior to Qwen2.5-coder-instruct, and obtain very close performance on other tests. 

\textbf{Model Scale Effects}
Experiments clearly show that the 32B model can better utilize high-quality synthetic data compared to the 7B model.On the 32B model, the average improvements compare to OSS-Instruct-GPT-4o is smaller (1-3\%), but this is still significant considering OSS-Instruct-GPT-4o already gets high-performance on benchmarks.The 7B model is more sensitive to data quality, with appropriate strategy combinations bringing over 10\% improvement.

\section{Conclusion}
Existing data generation methods for training large language models (LLMs) in code generation are constrained by high costs and limited diversity, hindering scalability and effectiveness. To address this, we propose Infinite-Instruct, a framework that automates high-quality instruction data synthesis through Reverse Construction and Backfeeding Construction , while leveraging cross-language static analysis to ensure data quality by filtering invalid samples.Our experiments show that Infinite-Instruct achieves 21.70\% performance gains on 7B models and 34.95\% on 32B models, with static analysis improving MBPP test accuracy from 74.80\% to 77.80\% on 7B models. These results demonstrate the scalability of automated data generation, the critical role of static analysis, and the ability to achieve state-of-the-art performance with less training data. 

\section{Limitations}
Despite the significant advantages our method demonstrates in generating code SFT data, several noteworthy limitations remain:
\begin{enumerate}
\item Simplistic Difficulty Adjustment Mechanism: The current method relies primarily on adjusting the number of keywords to increase problem complexity, which may lead to unpredictable complexity outcomes.
\item Inherent Limitations of Static Analysis: Static analysis techniques cannot comprehensively capture all types of code issues, particularly logical errors, algorithmic complexity problems, and architectural design flaws. These limitations constrain the upper bound of code quality improvement and may result in certain obscure yet serious problems being overlooked during the screening process.
\item Imbalanced Multi-language Support: Although our method supports multiple programming languages, the depth of support varies across different languages. Mainstream languages (such as Python and JavaScript) receive more comprehensive support, while some niche yet important domain-specific languages have weaker support.
\end{enumerate}
Future research will focus on four key directions: enhancing chain of thoughts generation to construct code question-answer pairs containing detailed reasoning steps; integrating dynamic and static analysis to comprehensively assess code quality; developing an adaptive difficulty tuning mechanism to match the needs of different proficiency models; and enhancing cross-language knowledge transfer for efficient multilingual data generation. These explorations will improve the quality and efficiency of code SFT data and lay a more solid foundation for the application of large-scale language models in the code domain.

\bibliography{custom}

\appendix
\section{Prompts}
\subsection{Integrated Reverse Question Generation Template}
\label{Integrated Reverse Question Generation Template}
\begin{lstlisting}[style=customMarkdown]
# Task
As a senior full-stack engineer, you need to design a high-quality [PROBLEM_TYPE] programming problem. You need to draw inspiration from input random code snippets to create problems that fit real-world scenarios.

# Characteristics of [PROBLEM_TYPE] Problems
## From the problem requirements:
1. [PROBLEM_TYPE specific requirement description]
2. [Common question patterns or keywords]
3. [Assessment focus]

## From the code requirements:
1. [Whether code must be provided]
2. [Code feature requirements]
3. [Relationship requirements with original code snippet]

# Workflow
1. Code Snippet Feature Analysis:
Analyze the programming language used, core functionality, implementation methods, technical characteristics and difficulty level, understand the code design thinking and application scenarios.

2. Inspiration Brainstorming:
Based on code characteristics, brainstorm [PROBLEM_TYPE related brainstorming focus].

3. Initial Problem Design:
Design the initial problem framework based on previous analysis and thinking, combined with # Characteristics of [PROBLEM_TYPE] Problems. The problem framework includes problem description and [related code requirements].

4. Problem Review and Optimization:
Review initial problems based on the following points:
   - [PROBLEM_TYPE specific review points]
   - Whether the code meets requirements
   - Whether the code is identical to the original code snippet, if so, it needs modification
   - Whether the problem difficulty/logic is appropriate

5. Problem Improvement and Revision:
Modify and improve based on issues found in review, optimize problem description and code.

6. Formal Problem Output: Strictly output problems according to the format below:
[Programming Language]: [Programming language that should be used for the answer]
[Problem Description]: [Describe your created problem in easy-to-understand language]

# Output Format
1. Code Snippet Feature Analysis:
2. Inspiration Brainstorming:
3. Initial Problem Design:
4. Problem Review and Optimization:
5. Problem Improvement and Revision:
6. Formal Problem Output: Strictly output problems according to the format below:
[Programming Language]: [Programming language that should be used for the answer]
[Problem Description]: [Describe your created problem in easy-to-understand language]

# Notes
1. Do not provide any solution ideas or hints in the problem.
2. Do not show any content in the problem that might suggest answers.
3. Problems should be written in clear and concise language that is easy to understand.
4. Written problems need to draw inspiration from input random code snippets but **cannot directly use the input random code snippets**.
5. [PROBLEM_TYPE specific additional notes]
\end{lstlisting}

When using the integrated template, replace the placeholders with the specific requirements for each problem type:

\subsubsection*{1. Code Generation}

\textbf{Characteristics from problem requirements:}
\begin{itemize}
    \item Problems must be solved by writing code
    \item Problems can have multiple solutions
    \item Keywords: implement, develop, build, write
\end{itemize}

\textbf{Characteristics from code requirements:}
\begin{itemize}
    \item No code should be provided in the problem
\end{itemize}

\textbf{Inspiration Brainstorming focus:}
\begin{itemize}
    \item Possible application scenarios and problem directions
    \item Different implementation angles
    \item Valuable problem points
\end{itemize}

\textbf{Problem Review additional points:}
\begin{itemize}
    \item Unreasonable: Directly using the input random code snippet as the answer to the problem
    \item Not complex enough: Problem requirements are too simple, lacking divergent thinking
\end{itemize}

\textbf{Additional notes:}
\begin{itemize}
    \item Problems must demonstrate being solved through code writing, but do not provide any form of code examples
\end{itemize}

\subsubsection*{2. Code Understanding}

\textbf{Characteristics from problem requirements:}
\begin{itemize}
    \item Typically starts with verbs such as "Explain", "Analyze", "Describe", "Clarify", etc.
    \item Sometimes requires explaining specific programming concepts or technical details
    \item Some questions may include the code's execution output, asking to explain the reasoning behind the output
\end{itemize}

\textbf{Characteristics from code requirements:}
\begin{itemize}
    \item Not allowed to be identical to the provided random code snippet
    \item Followed by a complete code block, which ideally contains multiple complete functions
\end{itemize}

\textbf{Inspiration Brainstorming focus:}
\begin{itemize}
    \item Potential application scenarios and problem directions
    \item Different implementation perspectives
    \item Valuable question points
\end{itemize}

\textbf{Problem Review additional points:}
\begin{itemize}
    \item Does the initial question framework provide the code to be explained?
    \item Whether the code to be explained is consistent with random code snippets
    \item Is the logic of the code to be explained too simple?
\end{itemize}

\subsubsection*{3. Knowledge-based Question}

\textbf{Characteristics from problem requirements:}
\begin{itemize}
    \item Use words like "explain", "analyze", "compare" to describe requirements
    \item Organize related concepts into coherent questions
    \item Each test point should focus on concept understanding
\end{itemize}

\textbf{Assessment Requirements:}
\begin{itemize}
    \item Conceptual accuracy
    \item Depth of principle understanding
    \item Practical application scenarios
    \item Pros and cons analysis
\end{itemize}

\textbf{Inspiration Brainstorming focus:}
\begin{itemize}
    \item Identify one core concept as the test point
\end{itemize}

\textbf{Problem Review additional points:}
\begin{itemize}
    \item Verify avoidance of code implementation tendency
    \item Whether it forms complete knowledge context
\end{itemize}

\subsubsection*{4. Code Completion}

\textbf{Characteristics from problem requirements:}
\begin{itemize}
    \item Words such as "complete", "fill in", "perfect", "supplement" will be used to describe the requirements
    \item Focuses on examining the ability to understand existing code structure and interfaces
    \item Maintains consistency by following code style and optimizes code structure to improve execution efficiency
\end{itemize}

\textbf{Characteristics from code requirements:}
\begin{itemize}
    \item Code must be provided where there are gaps in the logic
    \item There should be gaps in the code for completion
\end{itemize}

\textbf{Inspiration Brainstorming focus:}
\begin{itemize}
    \item Select a single and focused functional area
    \item Determine the difficulty level of the question
    \item Design the core algorithm or data structure
    \item Plan the location and scope of the code to be completed
\end{itemize}

\textbf{Problem Review additional points:}
\begin{itemize}
    \item Does the initial problem provide the code to be completed?
    \item Are there gaps in the code of the initial problem?
    \item Is the code to be completed consistent with the random code snippet?
    \item Is the logic of the code to be completed too simple?
\end{itemize}

\subsubsection*{5. Code Optimization}

\textbf{Characteristics from problem requirements:}
\begin{itemize}
    \item The problem should have multiple possible optimization directions
    \item Consider algorithm complexity, code structure, and implementation details
    \item Typically described using terms like "optimize", "improve", "refactor", etc.
    \item The problem should be concise and comprehensive
\end{itemize}

\textbf{Characteristics from code requirements:}
\begin{itemize}
    \item Not allowed to be identical to the provided random code snippet
    \item Should have clear efficiency, readability, or structural issues
\end{itemize}

\textbf{Inspiration Brainstorming focus:}
\begin{itemize}
    \item Design an initial code framework
    \item Identify multiple core optimization points to focus on
\end{itemize}

\textbf{Problem Review additional points:}
\begin{itemize}
    \item Does it provide code to be optimized?
    \item Is the code to be optimized identical to the provided random code snippet?
    \item Is the code optimization logic too simple?
\end{itemize}

\textbf{Additional notes:}
\begin{itemize}
    \item The final problem description output must include the code
\end{itemize}

\subsubsection*{6. Debug}

\textbf{Characteristics from problem requirements:}
\begin{itemize}
    \item The problem should contain multiple errors
    \item Problem sentence patterns may include:
    \begin{itemize}
        \item "The following code attempts to implement..."
        \item "An error occurs when running the following code..."
        \item "Please identify and fix the errors in the code..."
    \end{itemize}
\end{itemize}

\textbf{Characteristics from code requirements:}
\begin{itemize}
    \item The code may be incomplete or contain errors
    \item The code may include errors such as:
    \begin{itemize}
        \item Syntax errors
        \item Logical flaws
        \item Algorithm efficiency issues
        \item Boundary condition handling
        \item Incorrect use of data structure
    \end{itemize}
    \item The code to be fixed should not be identical to the provided random code snippet
\end{itemize}

\textbf{Inspiration Brainstorming focus:}
\begin{itemize}
    \item Design multiple error insertion strategies
\end{itemize}

\textbf{Problem Review additional points:}
\begin{itemize}
    \item Does it provide code to be fixed?
    \item Is the code to be fixed identical to the provided random code snippet?
    \item Does the code to be fixed contain multiple errors?
    \item Is the error logic too simple?
\end{itemize}

\textbf{Additional notes:}
\begin{itemize}
    \item Do not allow the content of possible problems to be solved
    \item Do not provide error point comments in the code to be fixed
\end{itemize}

\subsubsection*{7. Modify Code as required}

\textbf{Characteristics from problem requirements:}
\begin{itemize}
    \item The problem should have multiple modification requirements
    \item The problem typically uses verbs like "refactor" or "modify" etc, followed by specific requirements
\end{itemize}

\textbf{Characteristics from code requirements:}
\begin{itemize}
    \item The code to be modified must be provided, with clear functionality but space for optimization
\end{itemize}

\textbf{Inspiration Brainstorming focus:}
\begin{itemize}
    \item Design the core functionality of the initial code
    \item Plan multiple specific aspects that need modification
\end{itemize}

\textbf{Problem Review additional points:}
\begin{itemize}
    \item Does it provide the code to be modified?
    \item Is the code to be modified identical to the given random code snippet?
    \item Is the logic of the code to be modified too simple?
\end{itemize}

\textbf{Additional notes:}
\begin{itemize}
    \item Do not include modification comments in the code to be modified
    \item The final problem description output must include the code
\end{itemize}

\subsection{Complicate Prompt}
\label{Complicate Prompt}
\begin{lstlisting}[style=customMarkdown]
# Task
You will act as a prompt complexity expert, rewriting given prompts into more challenging versions that pose greater challenges to AI systems like ChatGPT. The rewrite must maintain human comprehensibility and executability.
# Complexity Methods (randomly select one)
1. Constraint Addition - Introduce additional restrictions or requirements
2. Depth Extension - Extend inquiry depth and expand scope
3. Concrete Specification - Replace abstract concepts with more specific expressions
4. Reasoning Refinement - Transform simple questions into forms requiring multi-step reasoning
5. Input Enhancement - Add data or code in specific formats, using question forms
6. Innovation Variation - Maintain domain and difficulty while creating more unique new prompts
# Workflow
1. Understanding Given Prompt: [Analyze theme, goals, difficulty, constraints, and domain]
2. Code Identification and Extraction: [Identify and fully extract all code blocks from original prompt, skip this step if original prompt contains no code]
3. Selected Method: [Choose appropriate complexity method based on understanding]
4. Selection Rationale: [Explain method selection rationale based on given prompt and chosen method]
5. Complexity Results:
   - Prompt Section: [Show complexified text content]
   - Code Section: [Insert extracted code blocks at original positions and formats, output "None" if none]
   - Completeness Verification: [Confirm all code blocks are correctly integrated]
# Output Format
1. Understanding Given Prompt:
2. Code Block Extraction:
3. Selected Method:
4. Selection Rationale:
5. Complexity Results:
   - Prompt Section:
   - Code Section:
   - Completeness Verification:
# Important Constraints
1. Code Completeness:
   - Fully preserve all code blocks, including language markers, indentation, line breaks, and comments
   - Maintain original positions of code blocks in complexity results
   - Prohibited from modifying code content and format
2. Prompt Requirements:
   - New content limited to 10-20 words
   - Ensure readability and executability
   - Special marker words prohibited
3. Task Boundaries:
   - Only complexify prompts, do not provide solution approaches
   - Use only one complexity method per time
\end{lstlisting}

\subsection{Text Rewrite Prompt}
\label{Text Rewrite Prompt}
\begin{lstlisting}[style=customMarkdown]
# Task
You are a text rewriting expert. Your task is to completely transform the **textual description** in the original question while preserving all non-textual elements.
All the output must be in English.

# Workflow
1. Paragraph Structure Analysis: (outline the abstract structure of paragraphs)
2. Logical Flow Analysis: (identify the function of each sentence and their logical relationships)
3. Paragraph Structure Brainstorming: (brainstorm 3 different approaches to restructure paragraphs using rewriting methods)
4. Sentence Structure Brainstorming: (brainstorm 3 different approaches to reconstruct sentences while maintaining question integrity)
5. Selected Approach: (specify which approaches from steps 3 and 4 you'll use)
6. Rewritten Question: (present the rewritten question while preserving all code, tables, and other non-textual elements)

# Output format
1. Paragraph Structure Analysis:
2. Logical Flow Analysis:
3. Paragraph Structure Brainstorming:
4. Sentence Structure Brainstorming:
5. Selected Approach:
6. Rewritten Question:

# Rewriting Methods
1. Alternative sentence ordering
2. Different syntactic structures (e.g., inversions, passives)
3. Various writing techniques (e.g., concise phrasing, strategic omission)
4. Alternative paragraph organization (e.g., purpose-first vs. background-first)

# Quality Standards
1. Avoid structural similarity - the rewritten version should differ significantly in paragraph and sentence patterns
2. Present only the final question without explanatory content
3. Maintain a neutral tone and professional, concise style without colloquialisms
4. Preserve all code, tables, and non-textual elements exactly as they appear
5. Exclude solution hints or guidance
6. Avoid content that might suggest answers
\end{lstlisting}

\subsection{Prompts for keyword extraction}
\label{Prompts for keyword extraction}
\subsubsection{Extract Task}
\begin{lstlisting}[style=customMarkdown]
# Task
Please deeply analyze the provided programming problems, extract the [Task] keywords from them, and summarize them in concise keyword form. You need to ensure the accuracy of the keywords.Output must be in English.

# Definition of Task
**[Task]** is defined as categorizing the main theme or operation of the input content into a broad domain or operation type for effective handling and application.

# Workflow
1. Input Reception: Receive and prepare to analyze input content.
2. Content Understanding: Carefully read the input content to grasp its main theme or operation.
3. Domain Categorization: Categorize the content into a broad domain or operation type.
4. Keyword Extraction: Describe this domain or operation type with a concise keyword.
5. Keyword Validation: Ensure the chosen keyword accurately reflects the core of the content, avoiding overly specific or detailed descriptions.
6. Result Formatting: Format and output the task type keyword according to the given output example.

# Output Format
[Task]:[keyword]

# Examples Given
Example 1:
Input:
{Given a string s containing just the characters '(', ')', '{', '}', '[', ']', determine if the input string is valid.

An input string is valid if:

Open brackets must be closed by the same type of brackets.
Open brackets must be closed in the correct order.
Every close bracket has a corresponding open bracket of the same type.}

Output:
[Task]:[String Validation]

Example 2:
Input:
{I have an employee payroll table with headers including employee ID, name, gender, age, department, daily wage, attendance days, and allowance. Please help me write a Python program to add a salary column to this table. (Salary = daily wage * attendance days + allowance)}

Output:
[Task]:[Data Processing]

Example 3:
Input:
{An additive number is a string whose digits can form an additive sequence.
A valid additive sequence must contain at least three numbers. Except for the first two numbers, each subsequent number in the sequence must be the sum of the two numbers before it.
Given a string containing only digits '0'-'9', write a Java algorithm to determine if the given input is an additive number. Return true if it is, and false otherwise.
Note: Numbers in the additive sequence cannot start with 0, except for the number 0 itself, so there won't be cases like 1, 2, 03 or 1, 02, 3.}

Output:
[Task]:[Sequence Validation]
\end{lstlisting}

\subsubsection{Extract Instruction}
\begin{lstlisting}[style=customMarkdown]
# Task
Please deeply analyze the provided programming problems, extract the [Instruction] keywords from them, and summarize them in concise keyword form. You need to ensure the accuracy of the keywords.Output must be in English.

# Definition of Instructions
1. [Instructions] are explicit requirements or constraints extracted from the input, used to guide task execution, output format, or operation steps.
2. **Instructions do not include programming languages specified in the problem**

# Workflow
1. Input Reception: Receive complete description of the programming problem.
2. Problem Understanding: Read the problem in detail to understand core requirements and objectives.
3. Instruction Identification: Carefully analyze the problem description to identify explicit [Instruction] content.
4. Instruction Extraction: Extract the identified instructions from the problem description.
5. Instruction Validation: Review the extracted instructions according to the definition of [Instructions], you need to ensure your output complies with the content in # Definition of Instructions.
6. Result Formatting: Organize and output the extracted instructions according to the given output example format.

# Output Format
[Instructions]:[keyword1] [keyword2]...

# Examples Given
Example 1:
Input:
{How to compare two lists in Python?}

Output:
[Instructions]:[]

Example 2:
Input:
{Please implement in Java: A company uses public telephone to transmit data, the data is a four-digit integer, which is encrypted during transmission,
The encryption rules are as follows: add 7 to each digit, then replace the digit with the remainder of the sum divided by 3, then swap the first and second digits, and swap the third and fourth digits.}

Output:
[Instructions]:[Implement digit encryption rules][Perform digit swapping]
\end{lstlisting}

\subsubsection{Extract Knowledge}
\begin{lstlisting}[style=customMarkdown]
# Task
Please deeply analyze the provided programming problems, extract the [Knowledge Points] keywords from them, and summarize them in concise keyword form. You need to ensure the accuracy of the keywords.Output must be in English.

# Definition of Knowledge Points
**[Knowledge Points]**: Core concepts, basic principles, or key operational steps necessary to solve specific programming problems, usually presented in the form of concise keywords.

# Workflow
1. Input Reception: Receive and prepare to analyze specific programming problems.
2. Problem Understanding: Read the problem thoroughly to grasp its core requirements and objectives.
3. Core Concept Identification: Identify basic theoretical knowledge and key programming principles needed to solve the problem.
4. Operation Step Extraction: Analyze main steps and necessary programming techniques for problem-solving.
5. Knowledge Point Condensation: Transform identified concepts and steps into concise keywords.
6. Key Point Validation: Review extracted knowledge points to ensure their necessity, completeness, and conciseness.
7. Result Formatting: Format and output key points according to the given output example.

# Output Format
[Knowledge Points]:[keyword1] [keyword2]...

# Examples Given
Example 1:
Input:
{Please implement in Java: A company uses public telephone to transmit data, the data is a four-digit integer, which is encrypted during transmission,
The encryption rules are as follows: add 7 to each digit, then replace the digit with the remainder of the sum divided by 3, then swap the first and second digits, and swap the third and fourth digits.}

Output:
[Knowledge Points]:[Modulo Operation][Number Processing][Data Swapping]

Example 2:
Input:
{Given an integer n, return the number of strings of length n that consist only of vowels (a, e, i, o, u) and are lexicographically sorted. A string s is lexicographically sorted if for all valid i, s[i] is the same as or comes before s[i+1] in the alphabet.}

Output:
[Knowledge Points]:[String Generation][Vowels][Lexicographical Order][Combination Counting]

Example 3:
Input:
{Please provide a solution in Java code for the following problem: There are some spherical balloons taped onto a flat wall that represents the XY plane. The balloons are represented as an integer array points where points[i] = [xstart, xend] denotes a balloon whose horizontal diameter stretches between xstart and xend. You do not know the exact y-coordinates of the balloons.
An arrow can be shot up exactly vertically from different points along the x-axis. A balloon with xstart and xend bursts by an arrow shot at x if $\text{xstart} \leq x \leq \text{xend}$. There is no limit to the number of arrows that can be shot. An arrow once shot keeps traveling up infinitely.
Given an array points, return the minimum number of arrows that must be shot to burst all balloons.}

Output:
[Knowledge Points]:[Array Operations][Conditional Logic][Mathematical Logic]
\end{lstlisting}

\subsection{Keyword combination and filtering}
\label{Keyword combination and filtering}
\subsubsection{Description of keyword types}
\begin{lstlisting}[style=customMarkdown]
# Task:
Extract information from user input and structure it into `Node` objects. Only output structured text, no code generation needed.Output must be in English.

# Entity Type Description:
Identify entities in the input and classify them into the following types:
- **Task**: High-level goals or topics that typically require multiple steps or operations to complete. Tasks are ultimate goals, such as "develop an e-commerce website" or "design a database system".
- **Knowledge Point**: Computer science knowledge required for tasks or instructions, such as "HTML", "Python" or "MySQL".
- **Instruction**: Specific functions, effects or operations that need to be implemented in the task, usually low-level, concrete steps. Instructions are specific means to accomplish tasks, such as "query optimization" or "table structure design".

# Node Extraction Rules:
1. Preserve all identified **Task** and **Knowledge Point** entities.
2. Only preserve **important instructions** directly related to the task's core objectives, remove minor or unimportant instructions.
3. Each `Node` object should include:
  - Unique identifier (`id`)
  - Entity type (`type`)

# Output Format:
- **Only output structured text** in the following format:
Node(id="entity name", type="entity type")
- **Do not generate code**, only output the `Node` list in text form.

# Work Steps:
1. Read through the input content, identify and classify entities.
2. Analyze logical relationships between entities to determine their categories.
3. Output extracted `Nodes` in specified format.

# Notes:
- **Important instructions**: Instructions directly related to task core objectives.
- **Unimportant instructions**: Instructions with minor or secondary impact on task core objectives.
- If there is no clear task topic in the input, analyze logical relationships between entities to infer appropriate task type.
- Task type should be unique and clear.

# Example:

**Input**:
[open website], [develop an e-commerce website], [HTML], [CSS], [JavaScript], [implement user registration], [shopping cart functionality], [mouse operations]
**Expected Output**:
Node(id="open website", type="instruction")
Node(id="develop an e-commerce website", type="task")
Node(id="HTML", type="knowledge point")
Node(id="CSS", type="knowledge point")
Node(id="JavaScript", type="knowledge point")
Node(id="implement user registration", type="instruction")
Node(id="shopping cart functionality", type="instruction")
Node(id="mouse operations", type="instruction")
\end{lstlisting}

\subsubsection{Relationship analyses}
\begin{lstlisting}[style=customMarkdown]
# Task:
Build relationship object triples from input nodes.Output must be in English.Here's what needs to be done:

# Relationship Extraction:
- You should identify relationships between Nodes extracted from the input content.
- Create a relationship object for each relationship.
- A relationship object should have a subject (subj) and an object (obj), which are Node objects representing the entities involved in the relationship.
- Each relationship should also have a type (type) and, where applicable, other attributes (such as weight, direction, etc.).

# Node Type and Relationship Type Mapping:
- **Instructions** and **Knowledge Points** may have a "displays" relationship.
- **Tasks** and **Instructions** may have a "requires" relationship.
- **Tasks** and **Knowledge Points** may have a "contains" relationship.
- If there is no logical connection between two nodes and they clearly belong to completely different domains, use an "unrelated" relationship.

# Relationship Building Process:
1. Parse and understand the id and type from input nodes.
2. Think deeply about the inherent connections between different nodes, combining computer knowledge.
3. Find relationships that exist between different nodes.
4. Mark as "unrelated" if no relationship exists between nodes and they clearly belong to completely different domains.

# Work Steps:
- Read through the provided content.
- Identify relationships between input Nodes.
- Provide extracted relationships in specified format.

# Output Format:
- Extracted relationships should be formatted as instances of the provided relationship class.
- Ensure extracted data conforms to the class definition structure.
- Your output format should be: subject (subj) + type + object (obj).

# Examples:

## Example 1:
Node:
Node(id="data visualization analysis", type='instruction')
Node(id='HTML line chart', type='knowledge point')
Node(id='business analysis report', type='task')

Expected Output:
data visualization analysis displays HTML line chart
business analysis report requires data visualization analysis
business analysis report contains HTML line chart

## Example 2:
Node:
Node(id="physical acceleration", type='task')
Node(id="navigation bar", type='instruction')

Expected Output:
physical acceleration unrelated navigation bar

## Example 3:
Node:
Node(id="user login", type='task')
Node(id="password verification", type='instruction')
Node(id="security", type='knowledge point')

Expected Output:
user login requires password verification
password verification contains security

## Example 4:
Node:
Node(id="artificial intelligence", type='task')
Node(id="psychology", type='knowledge point')
Node(id="ethics", type='knowledge point')

Expected Output:
artificial intelligence contains psychology
artificial intelligence contains ethics

## Example 5:
Node:
Node(id="quantum computing", type='task')
Node(id="artificial intelligence", type='task')
Node(id="blockchain", type='knowledge point')
Node(id="cryptography", type='knowledge point')

Expected Output:
quantum computing contains cryptography
artificial intelligence contains blockchain
quantum computing unrelated artificial intelligence
artificial intelligence unrelated cryptography

# Complex Relationship Handling:
- If multiple relationships exist between nodes, list each relationship separately.
- If nested relationships exist (where the subject or object in a relationship is itself a relationship), clearly mark them.
- If temporal relationships exist between nodes (such as "Task A" must be completed before "Task B"), note this in the relationship type.

# Error Handling:
- If input node format is incorrect, return error message: "Input node format incorrect".
- If unable to determine relationships between nodes, return: "Unable to determine relationship".
\end{lstlisting}

\subsubsection{Associative phrase construction}
\begin{lstlisting}[style=customMarkdown]
# Task Description:
Extract reasonable word combinations from user-input triples (subject + type + object). Reasonable word combinations refer to task-relevant keywords, phrases, or sentences that clearly express the core logic and relationships of the task.Output must be in English.

# Word Type Definitions:
1. **Task**: Categorized into broad domains or operation types based on the main theme or operation of the input. For example, "create game" is a task.
2. **Instruction**: Explicit requirements or constraints extracted from the input, used to guide task execution, output format, or operation steps. For example, "function to detect cookie value" is an instruction.
3. **Knowledge Point**: Core concepts, basic principles, or key operational steps needed to solve specific programming problems. For example, "using HTML, CSS, JavaScript" is a knowledge point.

# Rules for Extracting Reasonable Associated Word Combinations:
1. **Remove Irrelevant Words**: Delete words that have no semantic or logical relationship with other words (including tasks).
2. **Analyze Relationships**: Analyze the subject-object relationships between remaining words, clarifying their logical connections.
3. **Task Logic**: Organize relationships between words following the logic of "Task + Knowledge Point + Instruction", ensuring clear primary and secondary relationships.

# Output Format:
- Extracted reasonable word combinations should be formatted as instances of the provided reasonable word combination class.
- Ensure extracted data conforms to the class definition structure.
- Your output should be arranged according to the task logic or primary-secondary relationships between words.

# Working Steps:
1. **Read Input**: Carefully read input content, understand its core theme and logic.
2. **Determine Primary-Secondary Relationships**: Clarify primary-secondary relationships between input words, distinguish tasks, knowledge points, and instructions.
3. **Remove Irrelevant Words**: Delete words that have no semantic or logical relationship with other words.
4. **Organize Logic**: Organize relationships between words following the "Task + Knowledge Point + Instruction" logic.
5. **Format Output**: Output extracted reasonable word combinations according to specified format.

# Notes:
1. **Semantic Association**: Ensure extracted word combinations have clear semantic associations.
2. **Clear Logic**: Output content should have clear logic and distinct primary-secondary relationships.
3. **Standard Format**: Output strictly according to specified format, ensuring consistent data structure.

# Examples:

## Example 1:
create game contains create interface
create game needs function to detect cookie value
create game needs record time
create game needs ensure clear code structure
create game based on using HTML, CSS, JavaScript
create interface needs function to detect cookie value
create interface needs record time
create interface needs ensure clear code structure
create interface based on using HTML, CSS, JavaScript
create interface unrelated to physical acceleration

Expected output:
create game, using HTML, CSS, JavaScript, create interface, function to detect cookie value, record time, ensure clear code structure

## Example 2:
programming parameter definition unrelated to print character
programming parameter definition unrelated to default primary key field
handle missing values unrelated to print character
handle missing values unrelated to default primary key field
train test set unrelated to print character
train test set unrelated to default primary key field
investigate outliers unrelated to print character
investigate outliers unrelated to default primary key field
data analysis unrelated to print character
data analysis unrelated to default primary key field
problem solving unrelated to print character
problem solving unrelated to default primary key field
data splitting unrelated to print character
data splitting unrelated to default primary key field

Expected output:
No relevance
\end{lstlisting}

\subsection{Integrated Backfeeding Question Generation Template}
\label{Integrated Backfeeding Question Generation Template}
\begin{lstlisting}[style=customMarkdown]
# Role
As an examiner specialized in designing [QUESTION_TYPE] programming questions, your task is to create a high-quality question based on keywords provided by users. These questions should [QUESTION_PURPOSE]. Output must be in English.

# Characteristics of [QUESTION_TYPE] Questions
## From question requirements:
1. [QUESTION_TYPE specific requirement description]
2. [Common question patterns or keywords]
3. [Assessment focus]

## From code requirements:
1. [Whether code must be provided - specific requirements]
2. [Code feature requirements]

# Output Example
1. Consider logical relationships between keywords: [List meanings of keywords and their logical relationships]
2. Understand question characteristics: [Analyze characteristics of [QUESTION_TYPE] questions, provide elements you think must be included]
3. Consider how to organize keywords into questions: [Through brainstorming, think divergently about how keywords can work together to form a programming question]
4. Output initial question: [Combine above thoughts to propose initial question]
5. Review initial question: [Identify unreasonable or specific areas for improvement in initial question and propose modification examples]
6. Propose new question: [Fix question based on modification suggestions]
7. Repeat above steps, review and modify again until question meets requirements
8. Final question output: [Output final question without any guiding words (like "Question:") or any symbols]

# Question Standards
1. Can hide emphasis on which programming language to use, letting students derive related knowledge themselves
2. Please strictly follow the format in # Output Example to give your thinking process for each step, but don't directly output the content in [], and the last step must be the final question output
3. Don't provide any solution ideas or hints
4. Don't show any content that might suggest answers
5. [QUESTION_TYPE specific additional standards]
\end{lstlisting}

\subsection{Prompt Filtering}
\label{Prompt Filtering}
\begin{lstlisting}[style=customMarkdown]
# Task
Your task is to evaluate how input prompts enhance the capabilities of advanced AI assistants. For each input prompt, analyze it according to the following 7 criteria.

# Standards
1. Specificity: Does the prompt request specific, clear outputs without ambiguity? This allows AI to demonstrate its ability to follow instructions and generate precise, targeted responses.
2. Domain Knowledge: Does the prompt test AI's knowledge and understanding in specific domain(s)? The prompt must require strong prior knowledge or mastery of domain-specific concepts, theories, or principles.
3. Complexity: Does the prompt contain multiple components, variables, or depth and nuance? This evaluates AI's ability to handle complex, multifaceted problems beyond simple queries.
4. Problem-Solving: Does the prompt require active problem-solving: analyzing and clearly defining problems, then systematically developing and implementing solutions? Note that active problem-solving goes beyond reciting facts or following fixed instruction sets.
5. Creativity: Does the prompt require creative approaches or solutions? This tests AI's ability to generate novel ideas tailored to the specific needs of the request or current problem.
6. Technical Accuracy: Does the prompt require answers with high technical accuracy, correctness, and precision? This evaluates the reliability and truthfulness of AI outputs.
7. Real-World Application: Does the prompt relate to real-world applications? This tests AI's ability to provide practical information that can be implemented in real-life scenarios.

# Output Example
Evaluation Process: (Ensure explanation before determining if input meets each criterion)
Standards Met: (List standard numbers met in Python array format, e.g., [1, 2, 4, 6, 7])
\end{lstlisting}

\subsection{Instruction Complexity Assessment}
\label{Instruction Complexity Assessment}
\begin{lstlisting}[style=customMarkdown]
You are an expert at evaluating the difficulty of programming questions. Your responsibility is to assess various types of questions, including QA, multiple-choice, debugging tasks, code explanations, and more. Your goal is to assign a difficulty score ranging from 1 (easiest) to 10 (most difficult).
Note that you are tasked with evaluating user-submitted programming questions rather than answering them

Steps:
Think and Understand: First, read and think carefully to ensure that you fully comprehend the question's intent. Focus on what the question is asking and what skills or knowledge are required to solve it.
Analysis: Based on your understanding, use the scoring criteria to assess the difficulty of the question. Consider factors such as the complexity of the task, the level of programming experience required, and whether specialized knowledge is needed.
Score: Assign a score between 1 and 10, reflecting the difficulty based on your analysis.

Scoring Criteria:
1 points - Very Easy
- Basic questions that programming beginners can easily answer.
- No specialized knowledge or prior programming experience is required.
- Typical tasks include:
    - Simple syntax corrections (e.g., missing semicolons or parentheses).
    - Basic input/output operations (e.g., printing "Hello World" or reading user input).
    - Basic variable assignments or arithmetic operations (e.g., assigning a value to a variable or adding two numbers).
    - Fixing a simple typo in a piece of code.
    - Simple logical conditions (e.g., writing an if-else statement).
    - Basic loops (e.g., a for-loop to iterate over an array).
    
2 points - Basic Programming Task
- Suited for beginners who have undergone a short learning period.
- Typical tasks include:
    - Arrays and basic list manipulations (e.g., accessing array elements, adding elements).
    - Elementary software configuration tasks (e.g., installing a library, setting up an IDE, configuring environment variables).
    - Writing basic functions that take input and return output.
    - Basic debugging, such as finding and fixing simple runtime errors.
    - Basic file I/O (e.g., reading from and writing to a file).
    - Writing functions that involve loops, conditionals, and data manipulation.
    
3 points - Common Programming Task
- Suitable for users with some programming experience.
- Typical tasks include:
    - Basic use of common data structures like lists or dictionaries.
    - Simple algorithms like sorting (e.g., bubble sort) and linear search.
    - Software development tasks like basic database operations (e.g., inserting or querying data from a database).
    - Implementing basic math functions (e.g., finding the greatest common divisor).
    - Basic error handling (e.g., using try-catch blocks).
    - Introduction to object-oriented programming (e.g., creating classes and objects).
    
4 points - Entry-Level
- Suitable for developers just starting out in software development.
- Typical tasks include:
    - Data structures like linked lists, hash tables, stack and queue (e.g., implementing a singly linked list).
    - Algorithms like binary search, insertion sort.
    - Simple server-side programming (e.g., writing a basic HTTP server).
    - Designing and implementing basic APIs.
    - Debugging and testing small codebases (e.g., writing unit tests).

5 points - Lower Intermediate
- Suited for developers with 1-2 years of experience.
- Typical tasks include:
    - Complex algorithms (e.g., improving the time complexity of sorting from $O(n^2)$ to $O(n \log n)$).
    - Complex data structures (e.g., binary tree, heap).
    - Object-oriented programming with inheritance, polymorphism, and encapsulation (e.g., designing a class hierarchy).
    - Basic functional programming concepts (e.g., lambda expressions, higher-order functions).
    - Code debugging and performance optimization (e.g., optimizing a recursive function).
    - Development of small-scale systems, such as building a RESTful API or optimizing a database query.
    - Implementing simple design patterns (e.g., Singleton, Factory).
    - Using version control systems like Git for basic collaboration tasks.

6 points - Intermediate
- Suitable for developers with 3-4 years of experience.
- Typical tasks include:
    - Involvement with multi-module projects, such as writing modular and reusable code across different components.
    - More complex data algorithms like greedy and backtracking.
    - Performance optimizations (e.g., improving the time complexity of algorithms).
    - Designing and implementing moderately complex API interfaces (e.g., handling authentication and rate limiting).
    - Service integration (e.g., integrating a third-party API into a project).
    - Developing small-to-medium-sized system modules (e.g., creating a caching layer for an application).
    - Concurrency control in programming (e.g., handling race conditions in multi-threaded environments).
 
7 points - Upper Intermediate
- Suitable for developers with 5-6 years of experience.
- Typical tasks include:
    - Complex system designs, requiring architectural understanding of multi-tier applications.
    - Working with more complex data structures like balanced trees (e.g., AVL trees) and graphs (e.g., BFS, DFS) and algorithms (e.g. dynamic programming).
    - Tackling advanced multithreading and synchronization issues (e.g., handling deadlock in concurrent programming).
    - Distributed system design and implementation (e.g., designing a distributed file storage system).
    - Building and optimizing high-concurrency models (e.g., designing a system to handle millions of simultaneous requests).
    - Designing and implementing advanced networked applications (e.g., web crawlers).
    
8 points - Advanced
- Suitable for developers with 7-10 years of experience.
- Typical tasks include:
    - Advanced dynamic programming problems (e.g., solving longest common subsequence problems).
    - Complex graph algorithms (e.g., implementing Dijkstra's or A* algorithms).
    - Working with complex technical stacks that span multiple platforms and languages.
    - Solving distributed system challenges (e.g., ensuring data consistency across a distributed database).
    - Advanced performance optimization tasks (e.g., reducing latency in real-time systems).
    - Complex concurrency models and synchronization across multiple threads or processes.
    - System performance tuning at scale (e.g., profiling and optimizing system performance for millions of users).
    - Cross-domain integrations (e.g., integrating machine learning models into production-level systems).
    
9 points - Expert Level
- Suitable for developers with over 10 years of experience.
- Typical tasks include:
    - Designing and implementing domain-specific languages (DSLs).
    - Developing low-level hardware drivers (e.g., writing a device driver in C).
    - Real-time system design (e.g., building systems for high-frequency trading platforms).
    - Working with highly specialized technologies (e.g., cryptography, embedded systems).
    - Building complex, large-scale distributed systems that require deep expertise (e.g., designing a global load balancer for a cloud platform).
    - Handling complex issues in system architecture (e.g., data sharding and replication across geographically distributed servers).
    
10 points - Academic Research and Innovation
- Reserved for top-tier experts, typically in academia or cutting-edge research. These task usually requires deep expertise in system architecture and understanding of limitations in computing, as well as potentially coming up with an innovative or theoretical solution that goes beyond conventional programming.
- Typical tasks include:
    - Designing brand new algorithms (e.g., quantum computing algorithms).
    - Creating breakthrough solutions.
    - Solving complex challenges that require original, creative thinking and often involve interdisciplinary knowledge (e.g., combining AI with robotics to solve novel problems).
    - Research and development in bleeding-edge fields.
    - Leading complex projects that require significant technical and academic expertise (e.g., creating a new programming paradigm).

**Instruction Following**
   - Please adhere strictly to the provided output format in the few-shot examples.
   - Your response should consist of three essential sections: Thinking Steps, Analysis, Json Output.
\end{lstlisting}

\section{Static Analysis Examples}
\label{Static Analysis Examples}
\subsection{Python}

\textbf{Prompt}
\begin{lstlisting}[style=customMarkdown]
Extend the given function to support multiple regular expression patterns. Each pattern may involve various named groups, and multiple string lists should be processed. Each string across these lists must be parsed twice using different configurations. Successful matches should have their results compiled into dictionaries. Address handling of edge cases, specifically focusing on empty strings and non-matches, while verifying processing between different lists and pattern sets.
\end{lstlisting}

\noindent\textbf{Response}
\begin{lstlisting}[style=customMarkdown]
Certainly! Let's extend a given function to support parsing multiple lists of strings using multiple regular expression patterns. Each pattern will use named groups, and we will process each string twice using different configurations. Successful matches will be compiled into dictionaries. We will also address handling of edge cases like empty strings and non-matches.

Here's a Python implementation:
```Python
import re
from typing import List, Dict, Any

def parse_strings_with_patterns(patterns: List[str], string_lists: List[List[str]]) -> List[Dict[str, Any]]:
    # Compile all regex patterns beforehand for efficiency
    compiled_patterns = [re.compile(pattern) for pattern in patterns]

    # Storage for the parsed results
    results = []

    # Iterate over each list of strings
    for strings in string_lists:
        # Process each string
        for s in strings:
            # Skip empty strings
            if not s:
                continue

            # Prepare a dictionary to hold results for this particular string
            string_results = {'string': s, 'matches': []}

            # Parse each string using each pattern twice with different configurations
            for pattern in compiled_patterns:
                # First attempt: direct match
                direct_match = pattern.match(s)
                if direct_match:
                    # Save matched groups
                    string_results['matches'].append({
                        'config': 'direct_match',
                        'groups': direct_match.groupdict()
                    })

                # Second attempt: match all
                all_matches = pattern.findall(s)
                if all_matches:
                    # The result format of findall differs, we handle conversion here
                    if isinstance(all_matches[0], tuple):
                        for match in all_matches:
                            string_results['matches'].append({
                                'config': 'findall',
                                'groups': {k: v for k, v in zip(pattern.groupindex.keys(), match)}
                            })
                    else:
                        string_results['matches'].append({
                            'config': 'findall',
                            'groups': {list(pattern.groupindex.keys())[0]: match} for match in all_matches
                        })

            # Handle non-matches
            if not string_results['matches']:
                string_results['matches'].append({
                    'config': 'nomatch',
                    'groups': None
                })

            # Append the result for this string
            results.append(string_results)

    return results

# Example usage
patterns = [
    r'(?P<word>\w+)',
    r'(?P<number>\d+)',
]

string_lists = [
    ["Hello123", "Test456", ""],
    ["NoMatch", "789"]
]

results = parse_strings_with_patterns(patterns, string_lists)
for result in results:
    print(result)
```
### Explanation:

1. **Multiple Patterns**: We accept a list of patterns. Each pattern is expected to have named groups.

2. **Multiple String Lists**: We accept multiple lists, each containing strings to be parsed.

3. **Compiling Patterns**: All patterns are compiled up front for better performance.

4. **Processing Each String**: For each string in the lists:
   - Skip empty strings.
   - Attempt to match using each compiled pattern twice:
     - Direct match using `pattern.match()`.
     - Find all matches using `pattern.findall()` which returns tuples for multiple groups.
   - Extract named groups for successful matches and append them to the results.

5. **Handling Edge Cases**: 
   - Empty strings are skipped.
   - If no matches are found for a string, an entry is added with `config` set to `nomatch`.

6. **Compiling Results**: All successful matches and non-matches are collected in a structured format into a dictionary list that maintains all relevant information for further processing or analysis.

This extended function systematically processes a potentially complex set of inputs and configurations, providing a robust solution for varied parsing needs.
\end{lstlisting}

\noindent\textbf{Static analysis result}
\begin{lstlisting}[style=customjson,breaklines=true]
{
  "issues": [
    {
      "rule_name": "E0001:syntax-error", 
      "message": "Parsing failed: 'invalid syntax (current file, line 46)'",
      "position": {"start_line": 51, "end_line": 51}, 
      "severity": "error", 
      "content": "                            'groups': {list(pattern.groupindex.keys())[0]: match} for match in all_matches", 
      "language": "Python"
    }
  ], 
  "status": "fail"
}
\end{lstlisting}

\subsection{C/C++}

\textbf{Prompt}
\begin{lstlisting}[style=customMarkdown]
Create a thread-safe task manager class in C++. This class should be able to:

- Add new tasks with associated priority levels, while limiting the number of tasks running concurrently.
- Mark tasks as completed, ensuring that tasks with dependencies only proceed once dependencies are completed.
- Reset the completion status of all tasks.
- Guarantee multi-threaded safety and manage task execution in accordance with priority and dependency requirements.
\end{lstlisting}

\noindent\textbf{Response}
\begin{lstlisting}[style=customMarkdown]
To create a thread-safe task manager in C++, we will use several components from the C++ Standard Library, including `std::mutex`, `std::condition_variable`, and `std::thread`. We will use a priority queue to manage task priorities and a mechanism to track task dependencies. This example will focus on providing core functionalities and thread safety, but it's important to know that fleshing it out for production use might require additional error checking and optimization.

Here's a basic implementation:
```cpp
#include <iostream>
#include <queue>
#include <unordered_map>
#include <unordered_set>
#include <vector>
#include <thread>
#include <mutex>
#include <condition_variable>
#include <functional>
#include <atomic>

// Task structure
struct Task {
    int id;
    int priority;
    std::function<void()> execute;
    std::vector<int> dependencies;
    
    Task(int id, int priority, std::function<void()> f, std::vector<int> deps = {})
        : id(id), priority(priority), execute(std::move(f)), dependencies(std::move(deps)) {}
};

// Comparator for priority queue (higher priority runs first)
struct TaskCompare {
    bool operator()(const Task& t1, const Task& t2) {
        return t1.priority < t2.priority;
    }
};

class TaskManager {
public:
    TaskManager(int maxConcurrentTasks)
        : maxConcurrentTasks(maxConcurrentTasks), activeTasks(0) {}

    void addTask(int id, int priority, std::function<void()> taskFunction, std::vector<int> dependencies = {}) {
        std::unique_lock<std::mutex> lock(mutex);
        tasks[id] = Task(id, priority, std::move(taskFunction), std::move(dependencies));
        if (dependencies.empty()) {
            taskQueue.push(tasks[id]);
            cv.notify_one();
        }
    }

    void markTaskCompleted(int id) {
        std::unique_lock<std::mutex> lock(mutex);
        completedTasks.insert(id);
        for (auto& it : tasks) {
            Task& task = it.second;
            if (!completedTasks.count(task.id) && !waitingTasks.count(task.id)) {
                auto& deps = task.dependencies;
                if (std::all_of(deps.begin(), deps.end(), [this](int dep) { return completedTasks.count(dep); })) {
                    taskQueue.push(task);
                    waitingTasks.erase(task.id);
                    cv.notify_one();
                }
            }
        }
    }

    void resetCompletion() {
        std::unique_lock<std::mutex> lock(mutex);
        completedTasks.clear();
        while (!taskQueue.empty()) taskQueue.pop();
        waitingTasks.clear();
        for (auto& it : tasks) {
            if (it.second.dependencies.empty()) {
                taskQueue.push(it.second);
            } else {
                waitingTasks.insert(it.second.id);
            }
        }
        cv.notify_all();
    }

    void run() {
        std::vector<std::thread> threads;
        for (int i = 0; i < maxConcurrentTasks; ++i) {
            threads.emplace_back(&TaskManager::workerThread, this);
        }
        for (auto& thread : threads) {
            thread.join();
        }
    }

private:
    int maxConcurrentTasks;
    std::atomic<int> activeTasks;
    std::unordered_map<int, Task> tasks;
    std::unordered_set<int> completedTasks;
    std::unordered_set<int> waitingTasks;
    std::priority_queue<Task, std::vector<Task>, TaskCompare> taskQueue;
    std::mutex mutex;
    std::condition_variable cv;

    void workerThread() {
        while (true) {
            Task task(0, 0, []{}, {});
            {
                std::unique_lock<std::mutex> lock(mutex);
                cv.wait(lock, [this] { return !taskQueue.empty() || activeTasks > 0; });

                if (taskQueue.empty()) {
                    break;
                }
                
                task = taskQueue.top();
                taskQueue.pop();
                ++activeTasks;
            }

            // Execute the task
            task.execute();
            markTaskCompleted(task.id);

            {
                std::unique_lock<std::mutex> lock(mutex);
                --activeTasks;
                cv.notify_all();
            }
        }
    }
};

int main() {
    TaskManager tm(3); // Allow 3 tasks to run concurrently

    tm.addTask(1, 10, [] { std::cout << "Running Task 1\n"; });
    tm.addTask(2, 20, [] { std::cout << "Running Task 2\n"; });
    tm.addTask(3, 15, [] { std::cout << "Running Task 3\n"; });
    tm.addTask(4, 30, [] { std::cout << "Running Task 4\n"; }, {1, 2});
    tm.addTask(5, 25, [] { std::cout << "Running Task 5\n"; }, {3});

    // Run tasks, observing priorities and dependencies
    tm.run();

    return 0;
}
```

This implementation includes:

- **Task Structure**: Defines properties of a task, including its ID, priority, execution function, and dependencies.
- **Task Manager Class**: Manages tasks, constraints, and execution using a priority queue, mutex, and condition variables.
- **Add Task Method**: Adds tasks, specifying priorities and optional dependencies.
- **Mark Task Completed Method**: Marks tasks as completed and checks if dependent tasks can proceed.
- **Reset Method**: Resets all tasks' completion status and restarts execution.
- **Worker Threads**: Manage concurrent execution of tasks while observing their priorities and dependencies.

This setup allows for scalable and efficient task management in a multithreaded context, accommodating dependency resolution and priority-based execution. More features and error-handling might be necessary for a production system.
\end{lstlisting}

\noindent\textbf{Static analysis result}
\begin{lstlisting}[style=customjson,breaklines=true]
{
  "issues": [
    {
      "rule_name": "error",
      "message": "'all_of' is not a member of 'std'\n if (std::all_of(deps.begin(), deps.end(), [this](int dep) { return completedTasks.count(dep); })) {\n ^~~~~~",
      "position": {
        "start_line": 56,
        "end_line": 56
      },
      "severity": "error",
      "content": " if (std::all_of(deps.begin(), deps.end(), [this](int dep) { return completedTasks.count(dep); })) {",
      "language": "C++"
    },
    {
      "rule_name": "error",
      "message": "no matching function for call to 'Task::Task()'\n second(std::forward<_Args2>(std::get<_Indexes2>(__tuple2))...)\n ^",
      "position": {
        "start_line": 1673,
        "end_line": 1673
      },
      "file_path": "/usr/include/c++/8/tuple",
      "severity": "error",
      "language": "C++"
    }
  ],
  "status": "fail"
}
\end{lstlisting}

\subsection{Java}

\textbf{Prompt}
\begin{lstlisting}[style=customMarkdown]
Design a Java class tasked with executing and managing multiple tasks concurrently. Leverage the Builder pattern to configure the system, ensuring safe threading and efficient resource use. Support for executing SQL queries on specified databases should be included. The class must handle dependencies through priority levels, include comprehensive exception handling, and provide features for scheduling and canceling tasks. Additionally, define metrics to assess performance and consider the need for distributed task execution to address dynamic management challenges in real-world scenarios.
\end{lstlisting}

\noindent\textbf{Response}
\begin{lstlisting}[style=customMarkdown]
Creating a Java class that executes and manages multiple tasks concurrently using the builder pattern involves a structured approach to design not only for functionality but also for extensibility and maintainability. Here's a conceptual design of the class, including the necessary details such as exception handling, task dependencies, scheduling, and metrics for performance assessment.
```Java
import java.sql.Connection;
import java.sql.DriverManager;
import java.sql.ResultSet;
import java.sql.Statement;
import java.sql.SQLException;
import java.util.ArrayList;
import java.util.List;
import java.util.PriorityQueue;
import java.util.concurrent.Callable;
import java.util.concurrent.Executors;
import java.util.concurrent.ScheduledExecutorService;
import java.util.concurrent.Future;
import java.util.concurrent.TimeUnit;

// Define a Task class
class Task implements Comparable<Task> {
    private final int priority;
    private final Callable<?> callable;
    private final String description;

    public Task(int priority, Callable<?> callable, String description) {
        this.priority = priority;
        this.callable = callable;
        this.description = description;
    }

    public int getPriority() {
        return priority;
    }

    public Callable<?> getCallable() {
        return callable;
    }

    public String getDescription() {
        return description;
    }

    @Override
    public int compareTo(Task o) {
        return Integer.compare(o.getPriority(), this.priority);
    }
}

// Task manager class
public class TaskManager {
    private final ScheduledExecutorService executorService;
    private final PriorityQueue<Task> taskQueue;
    private final List<Future<?>> activeTasks;
    private final List<Exception> exceptions;

    // Private constructor
    private TaskManager(ScheduledExecutorService executorService, PriorityQueue<Task> taskQueue) {
        this.executorService = executorService;
        this.taskQueue = taskQueue;
        this.activeTasks = new ArrayList<>();
        this.exceptions = new ArrayList<>();
    }

    public static class Builder {
        private int numThreads;
        private final PriorityQueue<Task> taskQueue = new PriorityQueue<>();

        public Builder setNumThreads(int numThreads) {
            this.numThreads = numThreads;
            return this;
        }

        public Builder addTask(Task task) {
            taskQueue.add(task);
            return this;
        }

        public TaskManager build() {
            ScheduledExecutorService executorService = Executors.newScheduledThreadPool(numThreads);
            return new TaskManager(executorService, taskQueue);
        }
    }

    public void execute() throws InterruptedException {
        while (!taskQueue.isEmpty()) {
            Task task = taskQueue.poll(); // get highest priority
            Future<?> future = executorService.submit(() -> {
                try {
                    return task.getCallable().call();
                } catch (Exception e) {
                    exceptions.add(e);
                    // Log or handle exception
                }
            });
            activeTasks.add(future);
        }
        executorService.shutdown();
        executorService.awaitTermination(Long.MAX_VALUE, TimeUnit.NANOSECONDS);
    }

    public void scheduleTask(Task task, long delay, TimeUnit timeUnit) {
        executorService.schedule(() -> {
            try {
                task.getCallable().call();
            } catch (Exception e) {
                exceptions.add(e);
            }
        }, delay, timeUnit);
    }

    public void cancelAllTasks() {
        for (Future<?> task : activeTasks) {
            task.cancel(true);
        }
    }

    public List<Exception> getExceptions() {
        return exceptions;
    }

    // SQL Task Support
    public static Callable<ResultSet> createSQLTask(String query, String url, String user, String password) {
        return () -> {
            try (Connection connection = DriverManager.getConnection(url, user, password);
                 Statement statement = connection.createStatement()) {
                return statement.executeQuery(query);
            } catch (SQLException e) {
                throw new RuntimeException("SQL Task Failed", e);
            }
        };
    }

    // Metrics to assess performance
    public static class Metrics {
        private final long startTime;
        private long endTime;
        private int totalTasks;
        private int completedTasks;

        public Metrics() {
            this.startTime = System.currentTimeMillis();
        }

        public void taskCompleted() {
            completedTasks++;
        }

        public void setTotalTasks(int totalTasks) {
            this.totalTasks = totalTasks;
        }

        public void end() {
            this.endTime = System.currentTimeMillis();
        }

        public long getTotalTime() {
            return endTime - startTime;
        }

        public int getCompletedTasks() {
            return completedTasks;
        }

        public int getTotalTasks() {
            return totalTasks;
        }

        public double getCompletionRate() {
            return (double) completedTasks / totalTasks * 100;
        }
    }
    
    // Considerations for Distributed Task Execution:
    // To expand this design to a distributed system, additional components and mechanisms should be
    // integrated, like:
    //  - Distributed Task Queues (e.g., using message brokers like Kafka, RabbitMQ).
    //  - Database for task tracking, retries, and fault tolerance.
    //  - Task execution engines that can scale beyond what is available in one JVM,
    //    potentially leveraging cloud services like AWS Lambda for serverless execution.
}
```

### Key Points:

1. **Concurrency Management**: Uses `ScheduledExecutorService` to handle concurrent task execution and scheduling.
   
2. **Task Queuing with Priority**: Utilizes a `PriorityQueue` to manage tasks based on priority levels.

3. **Builder Pattern**: Allows flexible configuration of `TaskManager` with a specified number of threads and tasks.

4. **SQL Execution**: Implements a method to create SQL tasks with a specified query and database connection details.

5. **Exception Handling**: Robust approach to capturing and managing exceptions during task execution.

6. **Performance Metrics**: Provides a mechanism to measure execution time, completion rate, and count of completed tasks.

7. **Task Scheduling and Cancellation**: Offers functionalities to schedule tasks with delays and cancel running tasks.

8. **Distributed Execution Considerations**: While the current design is for a single JVM, it mentions the necessities for scaling towards distributed execution environments.
\end{lstlisting}

\noindent\textbf{Static analysis result}
\begin{lstlisting}[style=customjson,breaklines=true]
{
 "issues": [
   {
     "rule_name": "error",
     "message": "no suitable method found for submit(()->{ try [...]; } })\n Future<?> future = executorService.submit(() -> {\n ^\n method ExecutorService.<T#1>submit(Callable<T#1>) is not applicable\n (cannot infer type-variable(s) T#1\n (argument mismatch; bad return type in lambda expression\n missing return value))\n method ExecutorService.<T#2>submit(Runnable,T#2) is not applicable\n (cannot infer type-variable(s) T#2\n (actual and formal argument lists differ in length))\n method ExecutorService.submit(Runnable) is not applicable\n (argument mismatch; bad return type in lambda expression\n unexpected return value)\n where T#1,T#2 are type-variables:\n T#1 extends Object declared in method <T#1>submit(Callable<T#1>)\n T#2 extends Object declared in method <T#2>submit(Runnable,T#2)\n2 errors",
     "position": {
       "start_line": 86,
       "end_line": 86
     },
     "severity": "error",
     "content": " Future<?> future = executorService.submit(() -> {",
     "language": "Java"
   }
 ],
 "status": "fail"
}
\end{lstlisting}

\subsection{JavaScript}

\textbf{Prompt}
\begin{lstlisting}[style=customMarkdown]
When enhancing the `isCompletelyDefined` function, ensure it evaluates nested functions within objects. Special attention should be given to recursive structures. Return a boolean true when the functional properties are validated properly.
\end{lstlisting}

\noindent\textbf{Response}
\begin{lstlisting}[style=customMarkdown]
Enhancing the `isCompletelyDefined` function to evaluate nested functions within objects can be done by recursively traversing the object, checking each property for undefined values, and ensuring that all functions are valid and properly defined. Special attention to recursive structures necessitates keeping track of objects we've already encountered to prevent infinite loops.

Here is a potential implementation in JavaScript:

```javascript
function isCompletelyDefined(obj, visited = new WeakSet()) {
    // Check for null or non-object values
    if (obj === null || typeof obj !== 'object') {
        return true;
    }

    // Use a WeakSet to track visited objects to handle recursive structures
    if (visited.has(obj)) {
        return true;  // Avoid infinite loops
    }
    visited.add(obj);

    // Iterate over all properties of the object
    for (let key in obj) {
        if (obj.hasOwnProperty(key)) {
            const value = obj[key];
        
            if (value === undefined) {
                // If a property is explicitly undefined, return false
                return false;
            } 
            else if (typeof value === 'function') {
                // If the property is a function, assume it's valid if it's defined
                // You may add more complex checks here if necessary
                try {
                    value();
                } catch (e) {
                    return false;  // If function execution results in error, consider it invalid
                }
            } 
            else if (typeof value === 'object') {
                // For nested objects, recursively call isCompletelyDefined
                if (!isCompletelyDefined(value, visited)) {
                    return false;
                }
            }
        }
    }

    return true;
}
```

### Key Considerations:
- **Recursive Traversal**: The function recursively checks each property of the object. If a property is an object, it ensures those properties are also completely defined.
- **Handling of Functions**: It's assumed that as long as a function can be called without causing errors, it's properly defined. If a function property needs more complex validation, you can update the logic inside the function check.
- **WeakSet for Visited Objects**: To handle recursive structures safely, the function uses a `WeakSet` to keep track of visited objects, preventing an infinite loop on recursive references, such as circular linked lists or structures.
- **Error Handling**: Functions are simply called to check if they can execute without error, providing a basic form of validation. Depending on your context, you may want to further enhance this part.

This implementation ensures the structure is evaluated in-depth, focuses on comprehensive validation, and maintains efficiency for large objects.
\end{lstlisting}

\noindent\textbf{Static analysis result}
\begin{lstlisting}[style=customjson,breaklines=true]
{
  "issues": [
    {
      "rule_name": "no-prototype-builtins",
      "message": "Do not access Object.prototype method 'hasOwnProperty' from target object.",
      "position": {
        "start_line": 20,
        "end_line": 20
      },
      "severity": "error",
      "content": " if (obj.hasOwnProperty(key)) {",
      "language": "JavaScript"
    }
  ],
  "status": "fail"
}
\end{lstlisting}

\subsection{SQL}

\textbf{Prompt}
\begin{lstlisting}[style=customMarkdown]
Design a comprehensive system for topic management and display in a community forum. Implement functionalities to retrieve user-created topics using efficient database management techniques. Join necessary tables to compile complex data, such as user profiles, while applying context-aware filters, like temporal scopes or regional differences. Present the resulting data as a list of dictionaries, each featuring 'title', 'creator_name', 'creation_date', and optionally 'region'. Ensure dynamic pagination to manage extensive topic volumes efficiently, prioritizing descending order by creation date and evaluate resource usage implications.
\end{lstlisting}

\noindent\textbf{Response}
\begin{lstlisting}[style=customMarkdown]
Designing a comprehensive system for topic management and display in a community forum involves a combination of database design, efficient query construction, and dynamic data presentation. Here's a proposed design and implementation strategy:

### **System Overview**

1. **Database Design:**
   - **Tables:**
     - **Users:** Stores user information.
       - Columns: `user_id`, `name`, `email`, `region`
     - **Topics:** Stores topics created by users.
       - Columns: `topic_id`, `user_id`, `title`, `creation_date`, `content`
     - **Regions:** (Optional) Stores information about regions.
       - Columns: `region_id`, `region_name`

2. **Relations:**
   - A **User** can create multiple **Topics** (one-to-many relationship).
   - A **User** belongs to one **Region** (optional, many-to-one relationship).

### **Functionality Implementation**

#### **Data Retrieval:**

1. **Database Connection:**
   - Use a robust database management system like PostgreSQL or MySQL.
   - Implement connection pooling to manage multiple concurrent accesses efficiently.

2. **Query Construction:**
   - Utilize SQL to retrieve topic data with necessary joins and filters.
   - Example Query:
     ```sql
     SELECT 
       t.title, 
       u.name as creator_name, 
       t.creation_date, 
       u.region
     FROM 
       Topics t
     JOIN 
       Users u ON t.user_id = u.user_id
     ORDER BY 
       t.creation_date DESC
     LIMIT 
       %s OFFSET %s;
     ```
   - Use placeholders for pagination (`LIMIT %s OFFSET %s`).

3. **Context-Aware Filtering:**
   - **Temporal Scope:** Add a date range condition (`WHERE t.creation_date BETWEEN %s AND %s`).
   - **Regional Differences:**
     - Filter by region if specified (`WHERE u.region = %s`).

4. **Dynamic Pagination:**
   - Implement pagination by calculating `LIMIT` and `OFFSET` based on page number and page size.
   - Provide API endpoints with parameters for `page` and `page_size`.

### **Response Structure:**

- **Data Format:**
  - Return data as a list of dictionaries.
  - Example:
    ```python
    [
      {
        "title": "How to Learn Python",
        "creator_name": "Alice",
        "creation_date": "2023-10-01",
        "region": "North America"  # Optional
      },
      {
        "title": "Javascript Tips",
        "creator_name": "Bob",
        "creation_date": "2023-10-02",
        "region": "Europe"  # Optional
      }
    ]
    ```

### **Efficiency and Resource Management:**

1. **Indexing:**
   - Index `creation_date` for faster order operations.
   - Index foreign keys (`user_id`) for quicker join operations.

2. **Connection Pooling:**
   - Use a connection pool library to manage database connections efficiently.
   - Libraries like SQLAlchemy (Python) provide robust connection pooling options.

3. **Caching Strategy:**
   - Implement caching for repetitive queries using tools like Redis or Memcached.
   - Consider caching popular queries (e.g., when paginating through the latest topics).

4. **Load Testing:**
   - Conduct load testing to ensure that the system can handle extensive volumes without performance degradation.
   - Tools like JMeter or Locust can simulate concurrent users.

5. **Monitoring:**
   - Monitor database performance using built-in tools or third-party solutions.
   - Analyze query execution times and plan optimizations accordingly.

This design balances efficient data retrieval, context-aware filtering, and effective resource management, ensuring a scalable and responsive community forum system.
\end{lstlisting}

\noindent\textbf{Static analysis result}
\begin{lstlisting}[style=customjson,breaklines=true]
{
 "issues": [
   {
     "rule_name": "PRS:",
     "message": "Line 12, Position 1: Found unparsable section: 'LIMIT \\n %s OFFSET %s'",
     "position": {
       "start_line": 41,
       "end_line": 42
     },
     "severity": "error",
     "content": "LIMIT \n %s OFFSET %s;",
     "language": "SQL"
   }
 ],
 "status": "fail"
}
\end{lstlisting}

\end{document}